\documentclass[10pt,twocolumn,letterpaper]{article}

\usepackage[pagenumbers]{cvpr} % To force page numbers, e.g. for an arXiv version

\usepackage[dvipsnames]{xcolor}

\definecolor{cvprblue}{rgb}{0.21,0.49,0.74}
\usepackage[pagebackref,breaklinks,colorlinks,citecolor=cvprblue]{hyperref}

\usepackage{graphicx}
\usepackage{amsmath}
\usepackage{amssymb}
\usepackage{booktabs}
\usepackage{bm}
\usepackage{adjustbox}
\usepackage{mathtools}
\usepackage{amsfonts}
\usepackage{array}
\usepackage{tikz}
\usepackage{ctable}
\usepackage{float}
\usepackage{enumitem}
\usepackage{dsfont}
\usepackage{xcolor}         % colors
\usepackage{dashrule}       % for table
\usepackage{makecell}       % for table
\usepackage{colortbl}       % for table
\usepackage{times}
\usepackage{multirow}
\usepackage{subcaption}

\allowdisplaybreaks

\definecolor{blueblack}{RGB}{0, 108, 173}
\definecolor{taborange}{RGB}{235, 127, 14}
\definecolor{tabgreen}{RGB}{30, 160, 30}
\definecolor{tabpurple}{RGB}{128, 103, 189}
\definecolor{tabred}{RGB}{214, 39, 40}

\definecolor{sol_light_blue}{RGB}{38, 139, 210}
\definecolor{sol_blue}{RGB}{38, 139, 210}
\definecolor{nord_blue}{RGB}{38, 139, 210}
\definecolor{sol_green}{RGB}{163, 190, 140}
\definecolor{sol_red}{RGB}{220, 50, 47}
\definecolor{nord_red}{RGB}{250, 190, 192}
\definecolor{nord_green}{RGB}{163, 190, 140}

\definecolor{beer_orange}{RGB}{242, 142, 28}

\definecolor{nordblack}{RGB}{46, 52, 64}
\definecolor{nordred}{RGB}{191, 97, 106}
\definecolor{magenta}{RGB}{215, 10, 185}
\definecolor{nordgreen}{RGB}{143, 170, 120}
\definecolor{nordblue}{RGB}{94, 129, 172}
\definecolor{nordpurple}{RGB}{180, 142, 160}

\usepackage[capitalize]{cleveref}
\crefname{section}{Sec.}{Secs.}
\Crefname{section}{Section}{Sections}
\Crefname{table}{Table}{Tables}
\crefname{table}{Tab.}{Tabs.}

\newcommand{\A}{\mathbf{A}}

\newcommand{\I}{\mathbf{I}}
\newcommand{\h}{\mathbf{h}}
\renewcommand{\H}{\mathbf{H}}

\newcommand{\z}{\mathbf{z}}

\newcommand{\x}{\mathbf{x}}
\newcommand{\y}{\mathbf{y}}

\newcommand{\w}{\mathbf{w}}

\newcommand{\Q}{\mathbf{Q}}
\newcommand{\K}{\mathbf{K}}

\renewcommand{\z}{\mathbf{z}}
\newcommand{\Z}{\mathbf{Z}}

\newcommand{\X}{\mathbf{X}}

\newcommand{\Y}{\mathbf{Y}}
\newcommand{\W}{\mathbf{W}}

\newcommand{\V}{\mathbf{V}}

\renewcommand{\P}{\mathbf{P}}

\renewcommand{\V}{\mathbf{V}}

\newcommand{\M}{\mathbf{M}}

\definecolor{cvprblue}{rgb}{0.21,0.49,0.74}
\usepackage[pagebackref,breaklinks,colorlinks,citecolor=cvprblue]{hyperref}

\def\paperID{13861} % *** Enter the Paper ID here
\def\confName{CVPR}
\def\confYear{2024}

\title{Convolutional Initialization for Data-Efficient Vision Transformers}

\author{Jianqiao Zheng$^*$ \and Xueqian Li \and Simon Lucey \and \\
The University of Adelaide\\
{\tt\small    \url{https://github.com/osiriszjq/impulse_init}     }
}

\begin{document}
\maketitle

\begin{abstract}
    Training vision transformer networks on small datasets poses challenges. 
    In contrast, convolutional neural networks (CNNs) can achieve state-of-the-art performance by leveraging their architectural inductive bias. 
    In this paper, we investigate whether this inductive bias can be reinterpreted as an initialization bias within a vision transformer network. 
    Our approach is motivated by the finding that random impulse filters can achieve almost comparable performance to learned filters in CNNs. 
    We introduce a novel initialization strategy for transformer networks that can achieve comparable performance to CNNs on small datasets while preserving its architectural flexibility.
    \vspace{-0.35cm}
\end{abstract}

{
  \renewcommand{\thefootnote}%
    {\fnsymbol{footnote}}
  \footnotetext[1]{
  Corresponding e-mail: jianqiao.zheng@adelaide.edu.au.}
}

\section{Introduction}
\label{sec:intro}

\begin{figure*}[t]
    \centering
    {\includegraphics[width=0.95\textwidth]{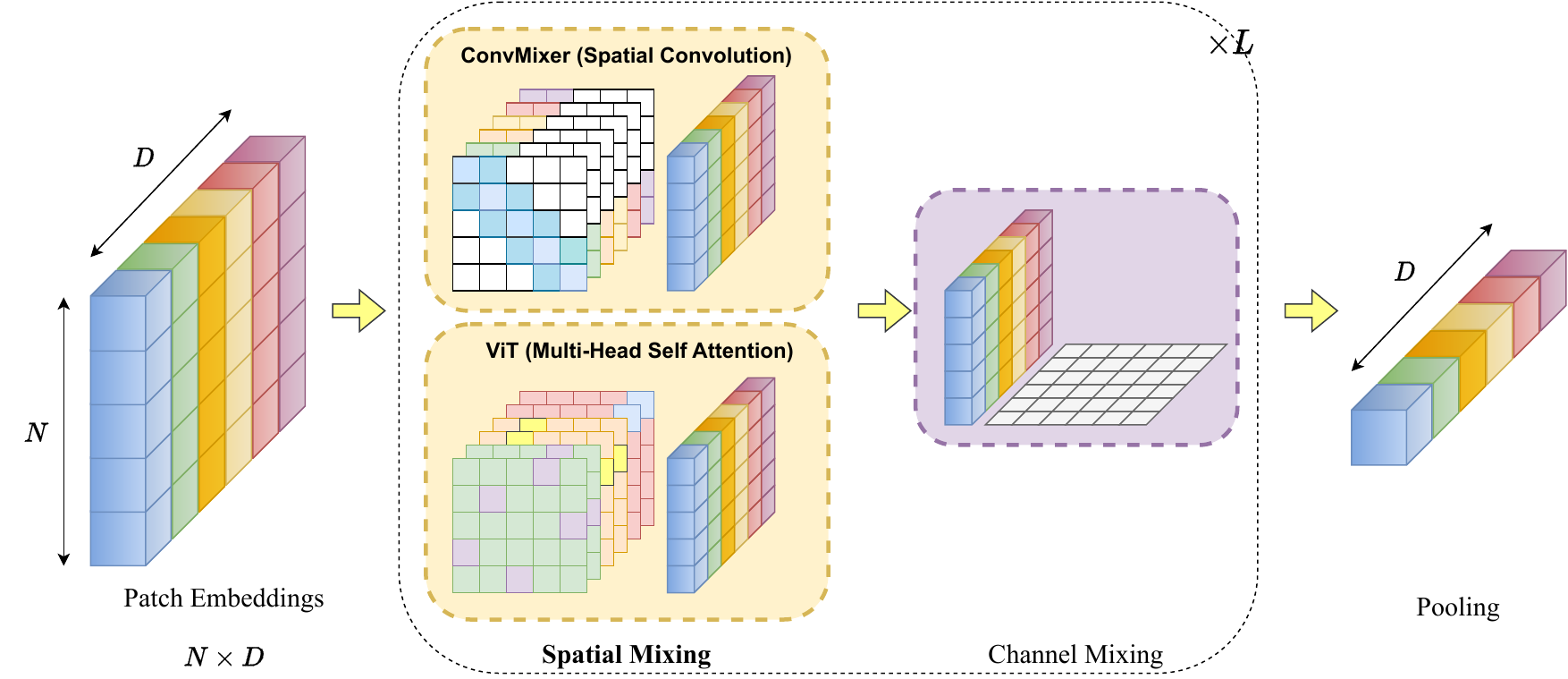} }%
    \caption{The architectures of ConvMixer~\cite{trockman2022patches} and Simple ViT~\cite{vit_baseline} are quite similar.
    Both are composed of input patch embedding, several layers of spatial mixing and channel mixing blocks, and then pooling for future fully connected classifiers.
    Skip connections, BatchNorm or LayerNorm, and ReLU or GeLU are not shown in this fig for simplification.
    The only difference is the structure of the spatial mixing matrix.
    In ConvMixer it is in convolution form (the upper one) and each channel has a different filter, while in Simple ViT (the lower one) the channels are divided into heads, and in each head, the spatial mixing matrix is the same which is computed by SoftMax of two low-rank matrices $\mathbf{Q}$ and $\mathbf{K}$.
    }%
    \vspace{-0.45cm}
    \label{fig:sim_conv_vit}%
\end{figure*}

Transformer networks have shown significant promise in vision problems when exposed to large amounts of data. 
However, when applied to smaller-scale datasets, their performance becomes poor compared more traditional convolutional neural networks (CNNs)~\cite{lee2021vision,liu2021efficient}. 
The superior performance of CNNs over visual transformers (ViTs) can largely be attributed to their convolutional inductive bias.

In addressing the drawback of training ViTs on small datasets, recent work~\cite{dosovitskiy2020vit} has employed pre-training on larger datasets such as ImageNet~\cite{ILSVRC15}, JFT-300M~\cite{sun2017revisiting},~\etc.
However, this approach has a fundamental limitation. 
One of the motivations for training visual learning algorithms, such as CNNs, on smaller datasets is the ability to quickly and cost-effectively evaluate their suitability for larger learning tasks. 
Requiring an algorithm to pre-train on a large dataset before training on a smaller dataset essentially undermines this purpose. 
Overcoming this limitation in modern ViTs could spark new waves of innovation and exploration in the vision community, making these models more accessible. 
This would enable researchers, educators, and students with limited computing resources to more readily investigate and explore the functions and applications of ViTs.

Recently, Trockman and Kolter~\cite{trockman2023mimetic} made noteworthy progress on this challenge.
They drew inspiration from the patterns visualized in the attention maps of ViTs pre-trained on large image datasets. 
Their approach requires no pre-training on large data and achieves comparable performance, allowing for easy adaptation of ViT architectures with little to no additional overhead. 
Despite its promising result, the approach is mimetic in that it attempts to mimic large-scale models without giving any theoretical understanding of how ViTs should be initialized. 
It also gives no connection to the convolutional inductive bias afforded by CNNs and how this potentially relates to data-efficient learning within ViTs.

Recent advances in CNNs---most notably ConvMixers~\cite{trockman2022patches}---have become increasingly similar in architecture to ViTs. 
Specifically, they break multi-channel convolution into depth-wise and channel-wise convolution. 
Unlike previous CNN variants, ConvMixers employ an equal resolution representation throughout all layers with no downsampling of the representation at successive layers. 
Transformer networks also employ channel-wise convolution, but replace the depth-wise convolution step with multi-headed attention (see~\cref{fig:sim_conv_vit}).

Cazenavette~\etal~\cite{cazenavetterethinking} recently demonstrated that random filter weights can exhibit comparable performance to learned weights in ConvMixer and ResNet~\cite{he2016deep} frameworks. 
The only parameters that need to be learned are those associated with channel-wise convolution, commonly referred to as channel mixing in the transformer literature~\cite{tolstikhin2021mlp}. 
This result is particularly intriguing, as the majority of parameters in a standard ConvMixer or ReseNet are largely redundant and do not require learning for effective performance. 
Expanding on this insight, we demonstrate that the utilization of random impulse filters achieves similarly impressive performance without the need for learning. 
In contrast to random filters, the convolutional matrix of a random impulse filter can be effectively modeled as a soft-max attention matrix, establishing a conceptual link between ConvMixer and ViT initialization. Specifically, we illustrate that the convolutional inductive bias inherent in CNNs through their architecture can also be realized through initialization within ViT.

\noindent Our paper makes the following contributions:-

\begin{itemize}
    \item We hypothesize that the multi-channel filters used within each layer of a CNN---such as those found in ConvMixer---should obey linear independence and redundancy. 
    Effective performance can be achieved solely by learning channel mixing when these two conditions are satisfied.
    \item Impulse filters satisfy these criteria and can be represented as a soft-max self-attention matrix, providing the core inspiration for our ViT initialization. 
    We evaluate this insight across a number of Simple ViT variants. 
    \item We demonstrate state-of-the-art performance for data-efficient ViT learning across several benchmarks including CIFAR-10, CIFAR-100~\cite{krizhevsky2009learning}, and SVHN~\cite{yuval2011reading}. 
    Our approach outperforms recent mimetic approaches and offers a more fundamental understanding of the ViT initialization.
\end{itemize}

\section{Related Work}
\label{sec:related_work}

\noindent \textbf{Convolution as attention:} 
Since their introduction~\cite{vaswani2017attention,dosovitskiy2020vit}, the relationship between transformers and CNNs has been a topic of immense interest to the vision community. 
Andreoli and Jean-Marc~\cite{andreoli2019convolution} studied the structure of attention and convolution, bringing them into a unified framework. 
Expanding on this groundwork, Cordonnier~\etal~\cite{Cordonnier2020On} notably demonstrated that self-attention layers can express any convolutional layers through a careful theoretical construction. 
Although promising, this earlier work mainly highlighted the equivalent functional capacity of self-attention in ViTs and convolutional spatial mixing in CNNs. 
It lacked insights into how the inductive bias of a ViT could be adapted or enhanced through this theoretical connection. 
Furthermore, the link between self-attention and convolution heavily relied on relative positional encoding---an element currently not widely used in mainstream ViT implementations.  
Our work, by contrast, leverages a simpler insight: random impulse filter convolution can be effectively approximated by soft-max self-attention.

\vspace{0.1cm}
\noindent \textbf{Bias through architecture:} 
Many efforts have attempted to embed convolutional inductive bias into ViTs through architecture modifications. 
Dai~\etal~\cite{dai2021coatnet} combined convolution and self-attention by mixing the convolutional self-attention layers. 
Both Pan~\etal~\cite{pan2021integration} and Li~\etal~\cite{li2023uniformer} proposed hybrid models of convolution and self-attention, where the output of each layer is a summation of the convolution and self-attention. 
Wu~\etal~\cite{wu2021cvt} explored the application of convolution as token projections within self-attention. 
Yuan~\etal~\cite{yuan2021incorporating} demonstrated promising results by inserting a depthwise convolution before the self-attention map as an alternate strategy for injecting inductive bias. 
All these techniques, however, have a fundamental limitation---they aim to introduce the inductive bias of convolution through architectural choices. 
Our work differs substantially by preserving the architectural flexibility of ViT through a novel CNN-inspired initialization strategy.
Such an approach is appealing to the current state-of-the-art as it can: (i) exhibit data efficiency on small datasets, (ii) have the architectural freedom to be seamlessly applied to larger datasets, and (iii) give an alternate theoretical perspective on how the inductive bias of convolution can be applied within transformers.

\vspace{0.1cm}
\noindent \textbf{Bias through initialization:} 
The application of inductive bias through initialization within a transformer has not been widely investigated to date. 
Zhang~\etal~\cite{zhang2022unveiling} posited that the benefit of pre-trained models in ViTs can be simply interpreted as a more effective strategy for initialization. 
Trockman~\etal~\cite{trockman2023mimetic,trockman2022understanding} recently studied the empirical distributions of self-attention weights---learned on large-scale datasets---and proposed their mimetic initialization strategy. 
A key difference in our work is that our method requires no off-line knowledge of pre-trained networks (mimetic or empirical). 
Instead, our strategy is intimately connected to the inductive bias of convolution, without having to resort to the introduction of convolution as an architectural choice.

\vspace{0.1cm}
\noindent \textbf{Simple ViT:} 
Many variants of ViT have been proposed since its introduction by Dosovitskiy~\etal~\cite{dosovitskiy2020vit}. 
Notably, several studies~\cite{lee2021vision, Gani_2022_BMVC,touvron2021training} 
have investigated how ViT can be adapted to achieve better performance on small-scale image datasets such as CIFAR-10~\cite{krizhevsky2009learning}. 
Of particular note in this regard is Simple ViT~\cite{vit_baseline}, which proposed several slight architectural modifications to ViT to improve performance on small-scale data, while maintaining good performance on large-scale tasks. 
As discussed in Section \ref{sec:method}, Simple ViT is attractive as its architectural similarity to ConvMixer~\cite{trockman2022patches} with the exception that the soft-max self-attention is replaced by conventional convolutional spatial mixing. 
Throughout this paper, we adopt Simple ViT as our baseline ViT model.

\section{Why Random Filters?}
\label{sec:sm_conv}
Cazenavette~\etal~\cite{cazenavetterethinking} recently demonstrated that randomly initialized convolution filters, in networks such as ConvMixer and ResNet, work remarkably well when only learning the channel mixing. 
However, they failed to provide any insight into the underlying reasons.
In this section, we offer a brief theoretical intuition. 
This result is significant as it builds a conceptual bridge between the architecture of ConvMixer and the initialization of ViT. 
For clarity and simplicity, we have omitted activations (\eg, GeLU, ReLU,~\etc), bias, batch norm, and skip connections in our formulation.

Let us define the patch embeddings in ConvMixer as $\X\,{=}\,[\x_{1},\x_{2},\dots,\x_{D}]$ where~$D$ is the number of channels and~$N$ is the number of pixels in the vectorized patch~$\x \,{\in}\, \mathcal{R}^{N}$. 
A 2D convolution filter is defined as~$\h \,{\in}\, \mathcal{R}^{f \,{\times}\, f}$, which can be represented as a convolutional matrix~$\H \,{\in}\, \mathcal{R}^{N \,{\times}\, N}$, such that~$\h \,{*}\, \x \,{=}\, \H \x$ where~$*$ is the convolutional operator. 
Let $\W \,{\in}\, \mathcal{R}^{D \,{\times}\, D}$ be the weight matrix of the channel mixing layer. 
Therefore, the result of spatial and channel mixing can be represented as,
\begin{equation}
    \Y = [\H_{1}\x_{1},\H_{2}\x_{2},\dots,\H_{D}\x_{D}]\W.
    \label{equ:spm_chm}
\end{equation}
Taking~$\Y \,{=}\, [\y_{1},\ldots,\y_{D}]$,~$\y \,{\in}\, \mathcal{R}^{N}$, one can define, 
\begin{equation}
    \y = \sum_{i=1}^D w_{i}\H_{i}\x_{i},
    \label{equ:spm_chm_02}
\end{equation}
where $\w \,{=}\, [w_{1},w_{2},\dots,w_{D}]^{T}$ is the corresponding vector slice of the learned channel mixing weights~$\mathbf{W} \,{=}\, [\w_{1},\ldots,\w_{D}]$.
Suppose the rank of $\X \,{=}\, \Z \A$ is $k$, and $k \,{\ll}\,D$, we obtain 
\begin{equation}
\begin{aligned}
    \y =\sum_{i=1}^D\sum_{j=1}^{k}  w_{i}a_{ji}\H_{i}\z_{j},
\end{aligned}
    \label{equ:spm_chm_z}
\end{equation}
where~$\Z \,{=}\, [\z_{1},\ldots,z_{k}]$ and~$a_{ji}$ refers to the row~$j$, column~$i$ element of~$\A$.

\begin{figure}[t]
    \centering
    {\includegraphics[width=1.0\linewidth]{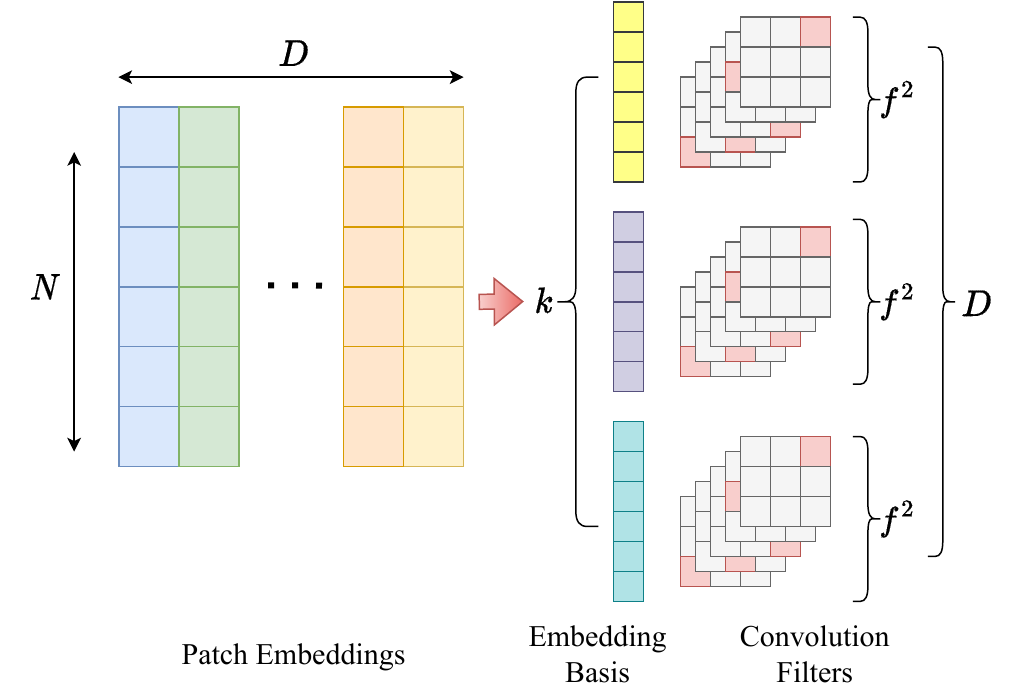} }%
    \caption{Illustration of why random spatial convolution filters are effective.
    }%
    \vspace{-0.3cm}
    \label{fig:sm_conv}%
\end{figure}

Remember that a linear combination of a~$f \,{\times}\, f$ linearly independent filters~$\h$ can express any arbitrary learned filter. 
Therefore, since $D$ is the number of channels, $k$ is the rank of input $\X$, and $f \,{\times}\, f$ is the number of convolution filters basis, we know that as long as
\begin{equation}
    D \geq k\cdot f^2,
    \label{equ:conv_condition}
\end{equation}
we can get any convolution result whose filters are in the space spanned by the given convolution filters by only learning the channel mixing parameters, as illustrated in~\cref{fig:sm_conv}. 
From~\cref{equ:spm_chm_z}, we see that $\y$ is just a linear combination of $k$ components $\z_{j}$, where each of the components is spatially filtered by an aggregation of $D$ filters. 
Even though this derivation omits activations, batch norms, and skip connections, it gives an intuition on how only learning channel mixing can be sufficient for achieving reasonably good performance. 
We back up these claims empirically in Section~\ref{sec:exp}.

\section{Method}
\label{sec:method}

\begin{figure*}[t]
    \centering
    {\includegraphics[width=0.95\linewidth]{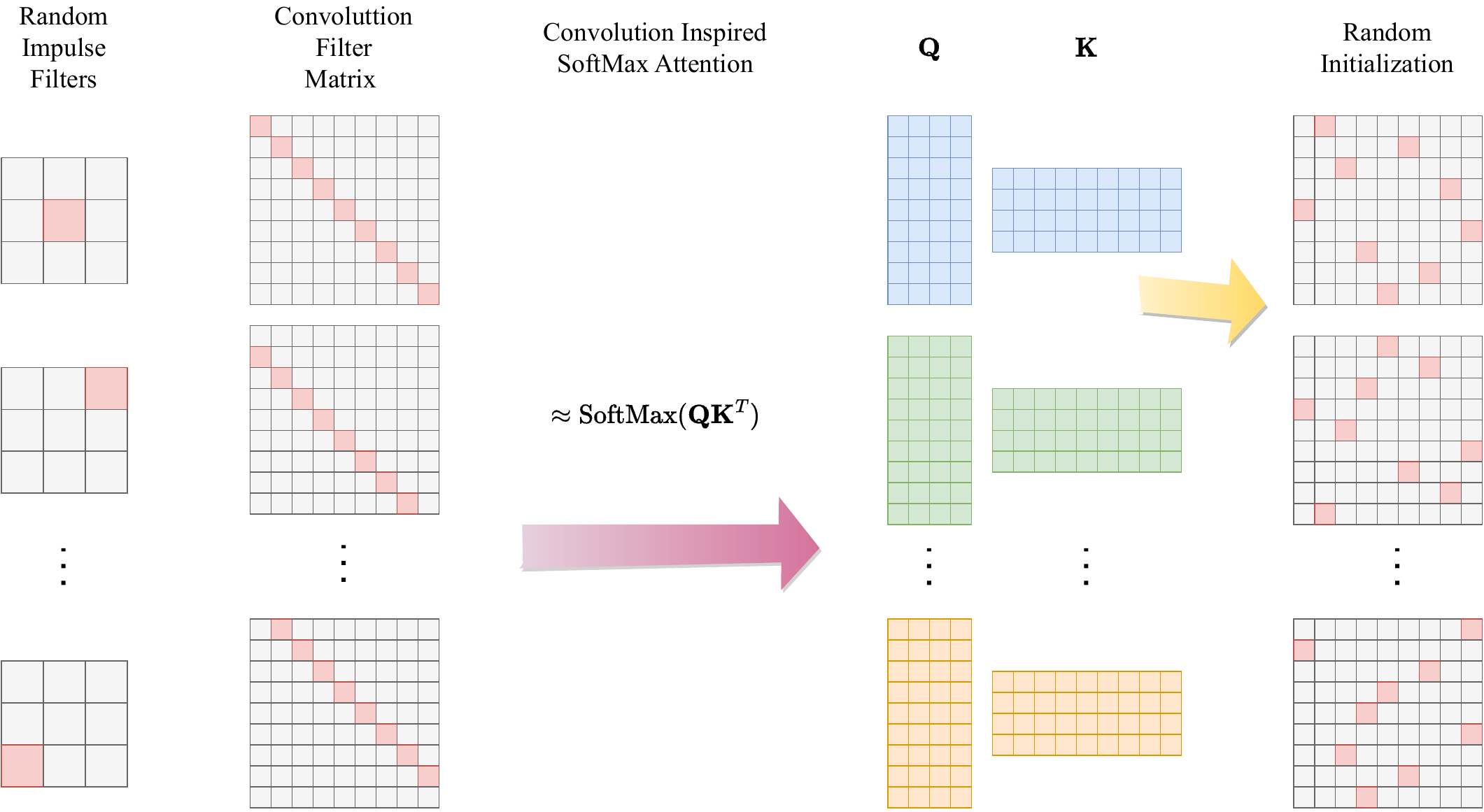} }%
    \caption{Our proposed strategy to initialize the weights of $\mathbf{Q}$ and $\mathbf{K}$ in the self-attention of transformers. 
    The pink cells indicate ones, while the gray cells are zeros. 
    Typically, the weights $\mathbf{Q}$ and $\mathbf{K}$ are randomly initialized, and after SoftMax, the attention map becomes a random permutation matrix, as indicated by the yellow arrow. 
    We propose to build convolution matrices of random impulse filters first and then initialize $\mathbf{Q}$ and $\mathbf{K}$ such that the initial attention map is a random impulse convolution filter, as shown by the purple arrow. 
    }%
    \vspace{-0.2cm}
    \label{fig:init}%
\end{figure*}

As shown in~\cref{fig:sim_conv_vit}, ConvMixer and ViT share most components in their architectures. 
The gap in their performance on small-scale datasets stems from their architectural choices of spatial mixing.
Although depthwise convolution (ConvMixer) and multi-head self-attention (ViT) may appear distinct at first glance, the goal of this step is the same. 
The spatial mixing step can be expressed as, 
\begin{equation}
    \mathbf{x} \leftarrow \M \mathbf{x},
\end{equation}
which are then followed by the channel mixing, and pooling steps as shown in~\cref{fig:init}.
~$\X \,{=}\, [\x_{1},\ldots,\x_{D}]$ are the patch embeddings where~$\x \,{\in}\, \mathcal{R}^{N}$ and~$\M$ is the~$N \,{\times}\, N$ spatial mixing matrix.

For ConvMixer, the matrix~$\M$ can be seen as the convolutional matrix~$\H$ in~\cref{equ:spm_chm}. 
For ViT, the matrix~$\M$ can be seen as
\begin{equation}
\M = \mbox{SoftMax}(\X \mathbf{Q} \mathbf{K}^{T} \mathbf{X}^{T}),
\end{equation}
where $\mathbf{Q},\mathbf{K} \,{\in}\, \mathcal{R}^{D \,{\times}\, K}$ are the attention weight matrices and~$K$ is the feature dimension in each head.

In both ConvMixer and Simple ViT, the spatial mixing matrix~$\M$ can vary for each column within~$\X$. 
In ConvMixer, a different mixing matrix (\ie, filter) is learned for each column vector within~$\X$. 
For ViT, the same mixing matrix may be used for a linear combination of the column vectors such that
\begin{equation}
    \mathbf{x} \leftarrow \M \mathbf{X} \mathbf{v},
\end{equation}
where~$\mathbf{v} \,{\in}\, \mathcal{R}^{D}$ is either a learnable or fixed vector that selects the linear combination within~$\X$ to be spatially mixed. 
The vectors~$\mathbf{v}$ that belong to the same mixing matrix (\ie, head) can be concatenated into a matrix~$\mathbf{V} \,{\in}\, \mathcal{R}^{D \,{\times}\, D/H}$ where~$H$ refers to the number of heads.

\vspace{0.1cm}
\noindent\textbf{Model I:} 
A common strategy within ViT is to augment the patch embeddings~$\X$ with an equally sized positional embedding~$\P$---such that $\X \,{\rightarrow}\, \X \,{+}\, \P$---to preserve the patch ordering within the transformer. 
In ViT, this positional encoding step is typically only applied at the first layer. 
For reasons that shall become clear, we shall also consider a strategy where positional encoding is directly applied within the self-attention mechanism of all layers, 
\begin{equation}
    \M_{\I} = \mbox{SoftMax}((\X+\P)\mathbf{Q}\mathbf{K}^{T}(\X+\P)^{T}) \;\;. 
    \label{equ:m1}
\end{equation}
However, in this situation, we do not include positional encoding anywhere else within the transformer network (\ie, $\X \,{\rightarrow}\, \X \,{+}\, \P$). 
In all our experiments, we notice almost no performance drop with this slight augmentation from traditional ViT. 
We shall refer to this variation of ViT as~\emph{Model I}.

\vspace{0.1cm}
\noindent\textbf{Model II:} 
Unlike ViT, the spatial mixing matrix within ConvMixer has no dependency upon the patch embeddings~$\X$. 
To build a connection between these two, we consider a soft-max self-attention mapping that is also independent of~$\X$. 
Here, we consider a spatial mixing attention map solely dependent on positional encoding as 
\begin{equation}
    \M_{\I\I} = \mbox{SoftMax}(\P \mathbf{Q} \mathbf{K}^{T} \P^{T}),
    \label{equ:m2}
\end{equation}
referred to herein as~\emph{Model II}.

\vspace{0.1cm}
\noindent\textbf{Model III:} 
ConvMixer also uses positional encoding~$\P$ in the formation of its mixing matrix. 
For completeness, we also consider a spatial attention mixing map where positional encoding has been entirely omitted:
\begin{equation}
    \M_{\I\I\I} = \mbox{SoftMax}(\hat{\mathbf{Q}}\hat{\mathbf{K}}^{T}),
    \label{equ:m3}
\end{equation}
where $\hat{\mathbf{Q}},\hat{\mathbf{K}}$ are now~$N \,{\times}\, K$ matrices not~$D \,{\times}\, K$. 
We refer to this as~\emph{Model III}.

\subsection{Impulse Filter Initialized ViT}
From~\cref{sec:sm_conv}, it is obvious that as long as the convolution filters have sufficient rank (\ie, linear independence) and there is enough redundancy in the number of channels, then only the channel mixing matrix~$\W$ needs to be learned. 
This implies that filters do not necessarily need to be randomly initialized, but can be any linearly independent set of filters. 
An interesting alternative to random filters in this regard is random impulse filters, as they satisfy the linear independence assumption, and their convolutional matrix can be well approximated by a softmax attention matrix.

This insight is crucial when it comes to understanding the poor performance of current ViTs when~$\mathbf{Q}$ and~$\mathbf{K}$ are initialized in the traditional manner (\ie, random). 
When passed through the softmax operator, this results in a random permutation matrix (see~\cref{fig:init,fig:sigma}). 
While this is beneficial for learning complicated attention maps, it becomes problematic when applied to smaller-scale learning tasks. 
However, by initializing $\mathbf{Q}$ and $\mathbf{K}$ to form a random impulse filter convolutional matrix, we can incorporate the implicit bias of a CNN without resorting to a more restrictive architecture.

\cref{fig:init} illustrates how we implement our initialization.
Note that we use \emph{Model I} for our ViT model instead of the original Simple ViT.
First, we initialize the number of heads of random impulse convolution filters.
Then, we convert the random impulse convolution filters into the convolutional matrix form, which is also an ideal initialization for the attention map.
We use these attention maps as targets and train the weights of $\mathbf{Q}$ and $\mathbf{K}$ in each head to minimize the MSE loss between $\mbox{SoftMax}(\P\mathbf{Q}\mathbf{K}^{T}\P^T)$ and the target attention maps.
This training just takes a few seconds, and after this, we use the trained $\Q$s and $\K$s as our initialization for \emph{Model I} Simple ViT.
Finally, the model is trained with input $\X$, which is $\mbox{SoftMax}((\X\,{+}\,\P)\mathbf{Q}\mathbf{K}^{T}(\X\,{+}\,\P)^T)$, for the classification task.

\section{Experiments}
\label{sec:exp}
\subsection{Spatial Mixing of ConvMixer}
\label{sec:exp_convmixer}
\begin{figure}[t]
    \centering
    {\includegraphics[width=0.95\linewidth]{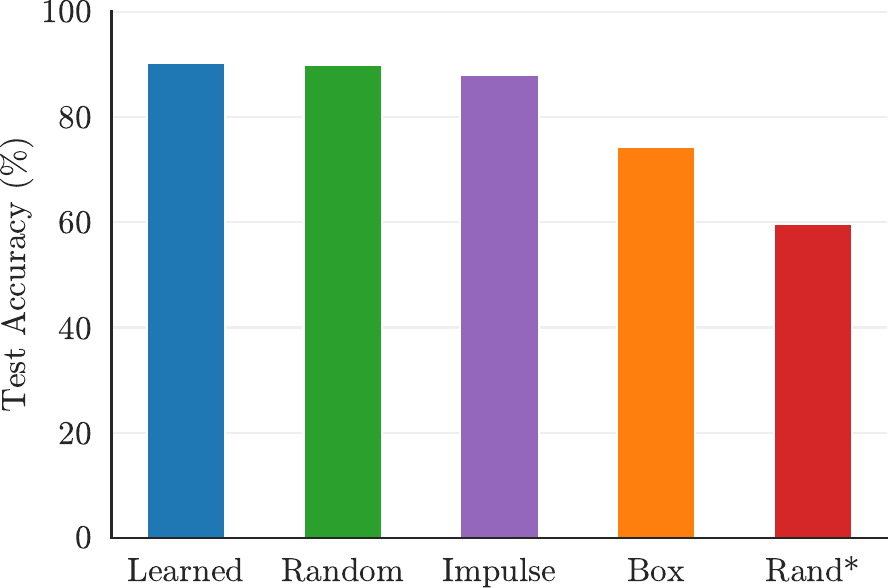} }%
    \caption{Performance comparison of models that use learned, random, impulse, box, and rand* (random permutation) filters. 
    Note that only the ``learned'' model has the spatial mixing weights that are trained, while all the other models have fixed weights.
    }%
    \vspace{-0.4cm}
    \label{fig:convmixer_base}%
\end{figure}

In~\cref{sec:sm_conv}, we provide theoretical insight into why randomly initialized filters within CNNs can achieve effective performance when only learning channel mixing. 
We propose that other filter choices, other than random, could be equally effective as long as they are linearly independent. 
One particularly interesting candidate in this context is random impulse filters. 
As shown in~\cref{fig:convmixer_base}, both random and random impulse filters perform approximately as well as the learned filters. 
For box filters (\ie, filter with all ones), despite redundancy in the number of channels, the filters are not linearly independent of each other and cannot span the entire filter space. 
As a consequence, the expressibility of the spatial mixing layer is reduced, resulting in the predicted worse performance. 
We also include results for a random perturbation matrix. 
This matrix is essentially an $N\,{\times}\,N$ matrix with a value one selected randomly in each row and zeros in all other positions. 
It differs from the random impulse filter matrix, as no convolutional constraint is given (see~\cref{fig:init}).

It also provides a valuable insight into why naively initialized self-attention matrices perform so poorly for small-scale learning tasks. 
When we randomly initialize weight matrices $\mathbf{Q}$ and $\mathbf{K}$ and then multiply them with $\mathbf{X}$, followed by SoftMax applied to each row, ones tend to appear at random positions with little consideration for locality. 
We argue that this may be a contributing factor to the challenges of training ViTs with small-scale datasets. 
Additional details on spatial mixing filters can be found in the appendix.

\subsection{Modified Simple ViT}
\label{sec:exp_modified_vit}
In this section, we evaluate three ViT variants, namely, \emph{Model I}, \emph{Model II}, and \emph{Model III}.
These experiments allow us to build a conceptual connection between ConvMixer and ViT. 
Our baseline model is the Simple ViT~\cite{vit_baseline} with specific configurations: embedding dimension of 512, 8 heads, depth of 6, and a patch size of 2.
For a fair comparison, we train all these models following the same settings asa described in~\cite{zhu2021vision}: a batch size of 512, the Adam optimizer with a learning rate of 0.0001, and training for 200 epochs. 
To truly evaluate their performance on small-scale data, we do not use any data augmentations in our baseline experiments since data augmentations will increase the scale of the data.
The input data is only normalized based on the mean and standard deviation of the data.

\begin{table}[t]
\caption[]{Classification accuracy(\%) on CIFAR-10 of modified Simple ViT. 
Specifically, for \emph{Model III}, ``R'' indicates random initialized weights in the format $\mbox{SoftMax}(\hat{\Q}\hat{\K}^{T})$, and ``T'' means these weights are trainable instead of fixed.}
\centering
\begin{adjustbox}{width=\linewidth}
\begin{tabular}{lcccccccc}
\toprule
\thead{\normalsize Model} & \thead{\normalsize Simple ViT} & \thead{\normalsize \emph{Model I}} & \thead{\normalsize \emph{Model II}}  & \thead{\normalsize \emph{Model III} (RT)} \\
\midrule
w/ value & 71.96 (57.97) & 71.3 & 72.16 & 73.04 \\
w/o value & 70.38 & 70.69 & 71.44 & 72.63 \\
\bottomrule
\end{tabular}
\label{tab:modified_vit}
\end{adjustbox}
\end{table}

\begin{table}[t]
\caption[]{Classification accuracy(\%) on CIFAR-10 of different initialization for \emph{Model III}: $\mbox{SoftMax}(\Q\K^{T})$. 
``T'' stands for trained weights and ``NT'' means the weights for queries and keys are not trained.
The latter implies that the spatial mixing operation is not trained.
``V'' refers to value.
{\color{tabgreen}\textbf{Green}} numbers indicate the percentage of accuracy increase with respect to the second-best performance.}
\centering
\begin{adjustbox}{width=\linewidth}
\begin{tabular}{@{}lcccccccc@{}}
\toprule
\thead{\normalsize Model} & \thead{\normalsize T w/ V} & \thead{\normalsize T w/o V} & \thead{\normalsize NT w/ V} & \thead{\normalsize NT w/o V} \\
\midrule
Random  & 73.04 $\hphantom{{\color{tabgreen}\textbf{2\%}}}$ 
        & \underline{72.63} $\hphantom{{\color{tabgreen}\textbf{2\%}}}$ 
        & \underline{61.76} $\hphantom{{\color{tabgreen}\textbf{26\%}}}$ 
        & 59.56 $\hphantom{{\color{tabgreen}\textbf{29\%}}}$ \\
Mimetic~\cite{trockman2023mimetic} 
        & \underline{73.46} $\hphantom{{\color{tabgreen}\textbf{2\%}}}$ 
        & 72.42 $\hphantom{{\color{tabgreen}\textbf{2\%}}}$ 
        & 61.24 $\hphantom{{\color{tabgreen}\textbf{26\%}}}$
        & \underline{59.98} $\hphantom{{\color{tabgreen}\textbf{29\%}}}$ \\
Impulse (Ours) 
        & \textbf{74.95} {\color{tabgreen}\textbf{2\%}} 
        & \textbf{73.94} {\color{tabgreen}\textbf{2\%}} 
        & \textbf{77.81} {\color{tabgreen}\textbf{26\%}} 
        & \textbf{77.11} {\color{tabgreen}\textbf{29\%}} \\
\bottomrule
\end{tabular}
\label{tab:intro_impulse}
\end{adjustbox}
\end{table}

\vspace{0.1cm}
\noindent\textbf{Modified models:} 
In~\cref{tab:modified_vit}, we present the basic results of our modified models on CIFAR-10~\cite{krizhevsky2009learning} without data augmentation. 
The baseline Simple ViT model achieves an accuracy of $71.91\%$.
However, it drops to $57.97\%$ when there is no input positional encoding, highlighting the significant effect of the positional information in ViT. 
In contrast, convolutional neural networks (CNNs) inherently capture positional information through their convolution mixing matrix.
Therefore, in common sense, there needs no explicit positional encodings.
This observation is supported by the comparison between ViT and CNNs shown in~\cref{fig:sim_conv_vit}, where spatial convolution corresponds to the attention map, necessitating positional information.

Building on this insight, \emph{Model I} introduces positional encoding only to the weights of $\mathbf{Q}$ and $\mathbf{K}$ in each layer, not the weights of $\mathbf{V}$. 
Despite this selective inclusion, the performance remains consistent at $71.3\%$,---which is similar to the baseline model---affirming our hypothesis. 
Changing to \emph{Model II} and \emph{Model III} that employ identity positional encoding, further increases test accuracy, aligning the results more closely with ConvMixer.

\begin{figure}[t]
    \centering
    \subfloat[\centering]{\includegraphics[width=0.25\linewidth]{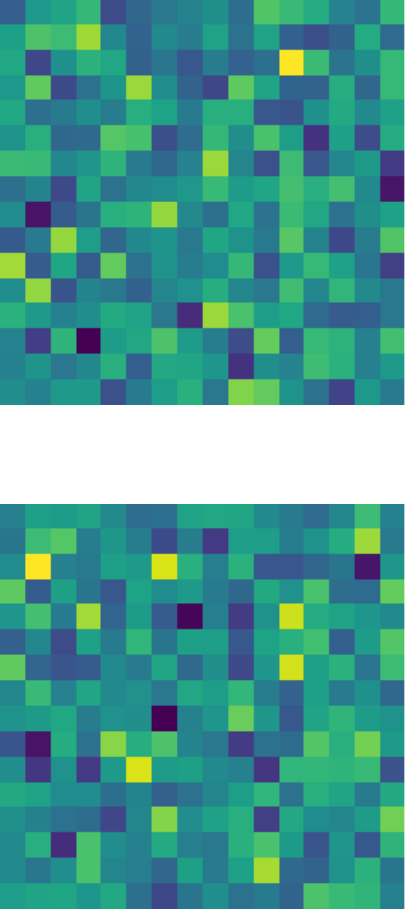}}%
    \qquad
    \subfloat[\centering]{\includegraphics[width=0.25\linewidth]{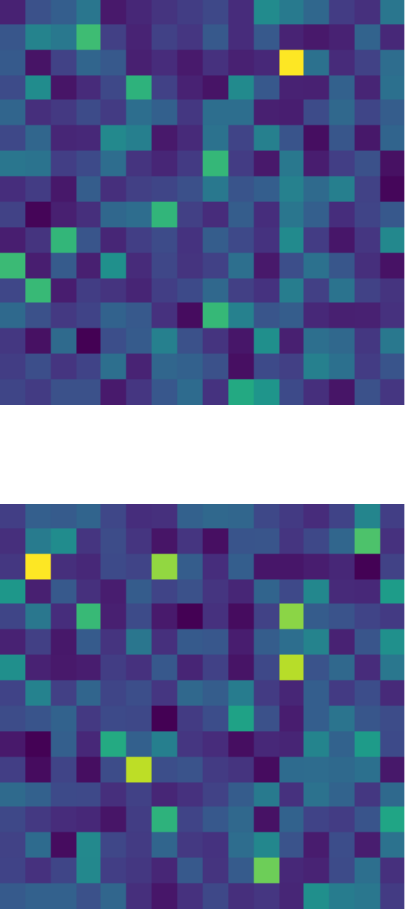}}%
    \qquad
    \subfloat[\centering]{\includegraphics[width=0.25\linewidth]{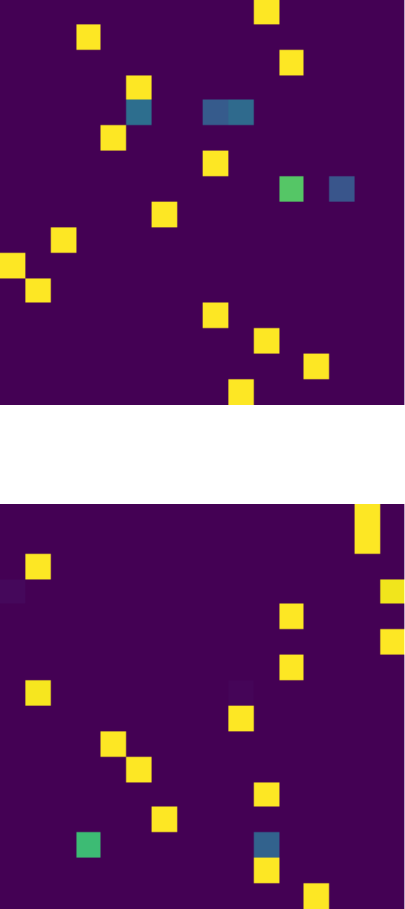}}%
    \caption{Illustration of how scale $\sigma$ affects the attention map. 
    In this $16 \,{\times}\, 16$ attention map example, $\M \,{=}\, \mbox{SoftMax}(\sigma\Q\K^{T})$, where $\Q$ and $\K$ are randomly initialized. 
    (a) $\sigma \,{=}\, 1.0$, (b) $\sigma \,{=}\, 1e2$ and (c) $\sigma \,{=}\, 1e4$.
    Larger sigmas tend to result in binary attention maps, which makes the attention map close to random permutation matrices. 
    This helps train the initial attention map to resemble the impulse convolution filter. 
    However, excessively large $sigma$ values will make the following training much more difficult.}%
    \vspace{-0.3cm}
    \label{fig:sigma}%
\end{figure}

\vspace{0.1cm}
\noindent\textbf{Impulse initialization:} 
In \emph{Model III}, the spatial mixing map is defined as $\M_{\I\I\I} \,{=}\, \mbox{SoftMax}(\Q\K^{T})$. 
As depicted in~\cref{fig:init}, we initialize several random $5 \,{\times}\, 5$ impulse filters, then convert them into a convolutional matrix of size $256 \,{\times}\, 256$, whose size equals the image size after patching. 
The resulting $\mbox{SoftMax}(\Q\K^{T})$ serves as a network with trainable parameters $\Q$ and $\K$. 
This process simulates the calculation of the attention map in self-attention layers. 

We train this small network to the expected random impulse convolution map with Mean Squared Error (MSE) loss, Adam optimizer, and a learning rate of $0.0001$. 
We train it for $10,000$ epochs, which converges well in a few seconds.
After this pre-training, we use these weights as the impulse initialization for the $\Q$ and $\K$ weights in self-attention. 
We also compare our approach to the mimetic initialization~\cite{trockman2023mimetic} and traditional random initialization, results shown in~\cref{tab:intro_impulse}. 
Excluding the input of $\X$ and $\P$ to the attention map, our impulse initialization outperforms the other two approaches on all different settings.

Specifically, our approach is slightly better (increases $2\%$ accuracy) than the other two approaches when using trained weights.
However, the performance of our approach outperforms the other two approaches by a large margin ($26\%$ and $29\%$) when using untrained weights with or without value $\V$.
Note that if the weights are randomly initialized and untrained, it is similar to the random permutation matrix shown in~\cref{fig:convmixer_base} while our impulse initialization is closer to the random impulse initialization in ConvMixer. 
However, minor model architectural differences (\eg, the number of heads in Simple ViT is 8, while it is 25 in ConvMixer) and training setting variations result in a performance gap between our impulse initialization and ConvMixer.

\vspace{0.1cm}
\noindent\textbf{Problem:} 
Note that in self-attention, the calculation of the attention map involves a constant $\sigma$ known as ``scale''. 
In reality, $\M_{\I\I\I} \,{=}\, \mbox{SoftMax}(\sigma\Q\K^{T})$. 
The original transformer~\cite{vaswani2017attention} introduces this scale to prevent the magnitude of the dot product from being too large and getting into the small gradient region of SoftMax.

Considering the properties of SoftMax, a small $\sigma$ tends to produce ``balanced'' attention maps with similar values, while a large $\sigma$ results in more ``pronounced'' maps with a lot of ones and zeros values (see~\cref{fig:sigma}). 
This $\sigma$ also controls how closely the initialized attention map can be trained to the impulse convolution filter. 
Consequently, choosing an appropriate $\sigma$ involves balancing these factors. 
In our experiments, we find that $\sigma \,{=}\, 1.0$ is an effective choice.

\begin{figure}[t]
    \centering
    {\includegraphics[width=0.95\linewidth]{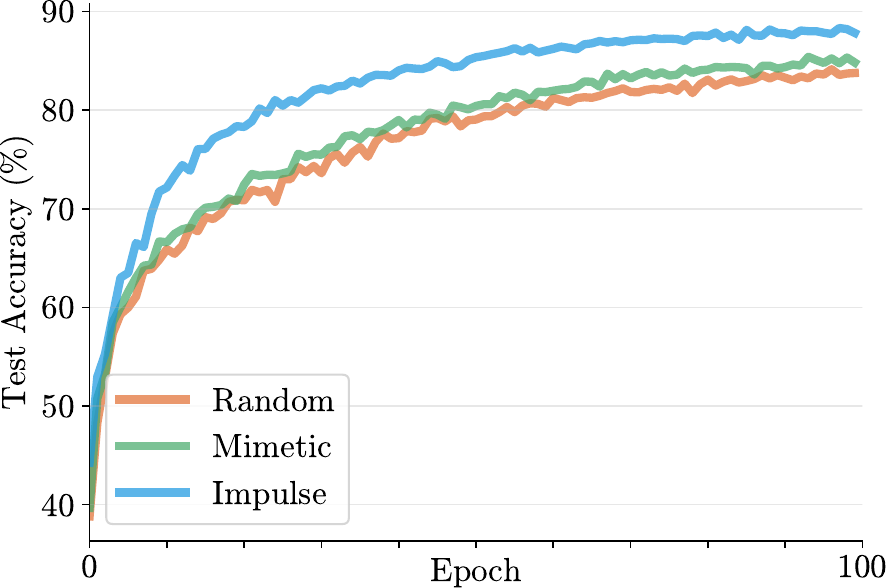} }%
    \caption{Training curves of three different initialization strategies on augmented CIFAR-10 with $\M'_{\I}$ for $\alpha \,{=}\, 0.2$. 
    Only the first 100 epochs are shown here.
    }%
    \vspace{-0.25cm}
    \label{fig:training_curve}%
\end{figure}

\subsection{Impulse Initialized Self-Attention}
\label{sec:exp_impulse_init}

In the last section, we discuss how to implement our impulse initialization approach \emph{Model III} defined as $\M_{\I\I\I} \,{=}\, \mbox{SoftMax}(\Q\K^{T})$. 
While this form allows the attention map to be a perfect impulse convolution filter, fixing it as an untrained map results in the loss of the ability to learn complex long-range relations. 
Training the attention map without the input $\X$ and $\P$ makes it challenging to learn long-range spatial relationships, potentially explaining why models with trained impulse filters perform worse than untrained ones, as observed in~\cref{tab:modified_vit}.
To retain the advantage of ViT for learning input-dependent attention maps and to incorporate the inductive bias of convolution, especially when dealing with small-scale datasets, we choose to initialize using $\M_{\I\I} \,{=}\, \mbox{SoftMax}(\P\Q\K^{T}\P^{T})$.
However, during training, we let the model operate in $\M_{\I} \,{=}\, \mbox{SoftMax}((\X \,{+}\, \P)\Q\K^{T}(\X \,{+}\, \P)^{T})$.

\begin{table}[t]
\caption[]{Classification accuracy(\%) on augmented CIFAR-10 of $\M'_{\I}$ in~\cref{equ:alpha_m1} with different initialization strategies and $\alpha$ values. 
When calculating the difference $\Delta$, ``R'' stands for random, ``M'' denotes mimetic, and ``I'' refers to our impulse initialization.
{\color{tabgreen}\textbf{Green}} numbers indicate the percentage of accuracy increase.}
\centering
\begin{adjustbox}{width=\linewidth}
\begin{tabular}{lcccccccc}
\toprule
\thead{\normalsize $\alpha$} & \thead{\normalsize 0.0} & \thead{\normalsize 0.1} & \thead{\normalsize 0.2} & \thead{\normalsize 0.3} & \thead{\normalsize 0.4} & \thead{\normalsize 0.5}\\
\midrule
Random & 87.26 & 86.87 & 85.95 & 84.19 & 83.56 & 82.53\\
Mimetic~\cite{trockman2023mimetic} & 87.63 & 88.05 & 86.88 & 86.77 & 86.26 & 86.05 \\
Impulse (Ours) & \textbf{90.01} & \textbf{90.45} & \textbf{89.97} & \textbf{89.53} & \textbf{88.92} & \textbf{88.35} \\
\midrule
$\Delta(\text{M}-\text{R})\;\%$ & {\color{tabgreen}0.42} & {\color{tabgreen}1.36} & {\color{tabgreen}1.08} & {\color{tabgreen}3.06} & {\color{tabgreen}3.23} & {\color{tabgreen}4.27} \\
$\Delta(\text{I}-\text{R})\;\%$ & {\color{tabgreen}\textbf{3.15}} & {\color{tabgreen}\textbf{4.12}} & {\color{tabgreen}\textbf{4.68}} & {\color{tabgreen}\textbf{6.34}} & {\color{tabgreen}\textbf{6.41}} & {\color{tabgreen}\textbf{7.05}} \\
\midrule
$\Delta(\text{I}-\text{M})\;\%$ & {\color{tabgreen}\textbf{2.72}} & {\color{tabgreen}\textbf{2.73}} & {\color{tabgreen}\textbf{3.56}} & {\color{tabgreen}\textbf{3.18}} & {\color{tabgreen}\textbf{3.08}} & {\color{tabgreen}\textbf{2.67}} \\
\bottomrule
\end{tabular}
\label{tab:cifar10_aug}
\end{adjustbox}
\end{table}

\vspace{0.1cm}
\noindent\textbf{Initialization:} 
The initialization step is similar to the steps described in~\cref{sec:exp_modified_vit}. 
Instead of training $\M_{\I\I\I} \,{=}\, \mbox{SoftMax}(\Q\K^{T})$ to be an impulse convolution filter, our goal now is to have $\M_{\I\I} \,{=}\, \mbox{SoftMax}(\P\Q\K^{T}\P^{T})$ represents random impulse convolution filters. 
Analogous to~\cref{sec:exp_modified_vit}, we build a small neural network to simulate the SoftMax and matrix multiplication of parameters $\Q$ and $\K$. 
Unlike~\cref{sec:exp_modified_vit}, this network takes positional encoding as input since it remains unchanged during the training process. 
Consistent with our previous settings, the neural network is trained to the expected random impulse convolution map with MSE loss, Adam optimizer, and a learning rate of $0.0001$ for $10000$ epochs.

\vspace{0.1cm}
\noindent\textbf{Tuning hyperparameters:} 
In our initialization of $\M_{\I\I}$, addressing the issue of $\sigma$ becomes crucial.
Training with $\P$ poses challenges as it is rank deficient, making it difficult to obtain the expected attention maps. 
During optimization, the model tends to push the magnitude of parameters to keep increasing, leading to parameters with large norms that hinder effective training. 
To solve this problem, we introduce weight normalization into this training process.
For the training of $\M_{\I\I}$ initialization, we empirically find that setting $\sigma \,{=}\, 0.1$ and $\eta \,{=}\, 100$ works well.
We adopt this configuration in all the following experiments for our impulse initialization approach.

\vspace{0.1cm}
\noindent\textbf{Implementation:} 
After we pre-train the initializations for $\Q$ and $\K$, we put these two weights into $\M_{\I} \,{=}\, \mbox{SoftMax}((\X \,{+}\, \P)\Q\K^{T}(\X \,{+}\, \P)^{T})$, and train for the classification task. 
In our implementation, we add a new parameter $\alpha$ to balance the input $\X$ and $\P$ in the form as
\begin{equation}
\begin{aligned}
    \M'_{\I} &= \mbox{SoftMax}\left(\Z\Q\K^{T}\Z^{T}\right),\\
    \Z&=\alpha\X+\left(1-\alpha\right)\P.
\end{aligned}
\label{equ:alpha_m1}
\end{equation}
When $\alpha \,{=}\, 0$, $\M'_{\I}$ becomes $\M_{\I\I}$ and when $\alpha \,{=}\, 0.5$, $\M'_{\I}$ becomes $\M_{\I}$.

\vspace{0.1cm}
\noindent\textbf{CIFAR-10:} 
We employ Rand Augment, Random Crop, and Random Horizontal Flip as data augmentations in our experiments.
\cref{tab:cifar10_aug} presents a comparison of three initialization strategies across different values of $\alpha$ from $0.0$ to $0.5$,~\ie, ranging from \emph{Model II} to \emph{Model I}, closely resembling the original Simple ViT. 
While mimetic initialization outperforms commonly used random initialization, our impulse initialization further boosts performance by $2.7\% \,{-}\, 3.6\%$, yielding a total performance gain of $3.2\% \,{-}\, 7.1\%$ compared to the random initialization. 
This highlights that initializing attention maps with random impulse convolution filters allows the model to leverage convolutional inductive bias while preserving the capacity to learn complex long-range relations.

An intriguing observation is that, with random initialization, decreasing $\alpha$ from $0.5$ to $0.0$ significantly boosts performance. 
This suggests that eliminating $\X$ from the input makes training easier. 
In scenarios where the dataset scale is limited, learning attention maps with information from $\X$ struggles to outperform data-independent attention maps. 
Convolutional inductive biased maps may provide the most effective prior knowledge in such cases.
The gap between random initialization and mimetic or impulse initialization is smaller at $\alpha \,{=}\, 0.0$, indicating that both designed initialization strategies, introducing inductive bias, help moderate the difference between data-dependent and data-independent attention maps. 
Another noteworthy point is that, for random attention maps, $\alpha \,{=}\, 0.0$ yields the best performance, as discussed earlier. 
However, for mimetic and our impulse filters, a sweet spot emerges at $\alpha \,{=}\, 0.1$, indicating that introducing a small amount of input $\X$ aids the model in learning more complicated relations based on the inductive bias.

\begin{table}[t]
\caption[]{Classification accuracy(\%) on augmented CIFAR-100 of $\M'_{\I}$ in~\cref{equ:alpha_m1} with different initialization strategies and $\alpha$ values.
When calculating the difference $\Delta$, ``R'' stands for random, ``M'' denotes mimetic, and ``I'' refers to our impulse initialization.
{\color{tabgreen}\textbf{Green}} numbers indicate the percentage of accuracy increase.}
\centering
\begin{adjustbox}{width=\linewidth}
\begin{tabular}{lcccccccc}
\toprule
\thead{\normalsize $\alpha$} & \thead{\normalsize 0.0} & \thead{\normalsize 0.1} & \thead{\normalsize 0.2} & \thead{\normalsize 0.3} & \thead{\normalsize 0.4} & \thead{\normalsize 0.5}\\
\midrule
Random & 61.90 & 64.16 & 62.52 & 61.07 & 59.47 & 57.98 \\
Mimetic~\cite{trockman2023mimetic} & 63.07 & 65.17 & 64.30 & 63.13 & 62.62 & 60.74 \\
Impulse (Ours) & 67.51 & 68.78 & 67.47 & 66.68 & 66.92 & 66.14\\
\midrule
$\Delta(\text{M}-\text{R})\;\%$ & {\color{tabgreen}1.89} & {\color{tabgreen}1.57} & {\color{tabgreen}2.85} & {\color{tabgreen}3.37} & {\color{tabgreen}5.30} & {\color{tabgreen}4.76} \\
$\Delta(\text{I}-\text{R})\;\%$ & {\color{tabgreen}\textbf{9.06}} & {\color{tabgreen}\textbf{7.20}} & {\color{tabgreen}\textbf{7.92}} & {\color{tabgreen}\textbf{9.19}} & {\color{tabgreen}\textbf{12.53}} & {\color{tabgreen}\textbf{14.07}} \\
\midrule
$\Delta(\text{I}-\text{M})\;\%$ & {\color{tabgreen}\textbf{7.04}} & {\color{tabgreen}\textbf{5.54}} & {\color{tabgreen}\textbf{4.93}} & {\color{tabgreen}\textbf{5.62}} & {\color{tabgreen}\textbf{6.87}} & {\color{tabgreen}\textbf{8.89}} \\
\bottomrule
\end{tabular}
\label{tab:cifar100_aug}
\end{adjustbox}
\end{table}

\vspace{0.1cm}
\noindent\textbf{Convergence speed:} 
In~\cref{fig:training_curve}, we show the training process of $\M'_{\I}$ for $\alpha \,{=}\, 0.2$ on augmented CIFAR-10 dataset. 
We only show the test accuracy in the first 100 epochs for clarity. 
Notably, mimetic initialization converges faster than random ones, while our impulse initialization converges significantly faster than the other two.

\vspace{0.1cm}
\noindent\textbf{Other datasets:} 
We extend the evaluation of different initialization strategies on CIFAR-100~\cite{krizhevsky2009learning} and SVHN~\cite{yuval2011reading}. 
All the model configurations and training settings remain consistent with previous experiments. 
Note that the SVHN dataset is evaluated without any data augmentation.
Results are presented in~\cref{tab:cifar100_aug} and~\cref{tab:svhn_noaug}, demonstrating the promising and consistent performance of our initialization approach, as observed in CIFAR-10 experiments.

\begin{table}[t]
\caption[]{Classification accuracy(\%) on no-augmented SVHN of $\M'_{\I}$ in~\cref{equ:alpha_m1} with different initialization strategies and $\alpha$ values. 
When calculating the difference $\Delta$, ``R'' stands for random, ``M'' denotes mimetic, and ``I'' refers to our impulse initialization.
{\color{tabgreen}\textbf{Green}} numbers indicate the percentage of accuracy increase, and {\color{tabred}\textbf{red}} numbers indicate the percentage of accuracy decrease.}
\centering
\begin{adjustbox}{width=\linewidth}
\begin{tabular}{lcccccccc}
\toprule
\thead{\normalsize $\alpha$} & \thead{\normalsize 0.0} & \thead{\normalsize 0.1} & \thead{\normalsize 0.2} & \thead{\normalsize 0.3} & \thead{\normalsize 0.4} & \thead{\normalsize 0.5}\\
\midrule
Random & 90.38 & 90.80 & 90.50 & 90.85 & 90.32 & 89.83 \\
Mimetic~\cite{trockman2023mimetic} & 90.12 & 91.35 & 92.03 & 91.63 & 91.25 & 90.47 \\
Impulse (Ours) & 91.96 & 93.37 & 93.85 & 93.33 & 93.44 & 92.96 \\
\midrule
$\Delta(\text{M}-\text{R})\;\%$ & {\color{tabred}-0.29} & {\color{tabgreen}0.61} &  {\color{tabgreen}1.69} & {\color{tabgreen}0.86} & {\color{tabgreen}1.03} & {\color{tabgreen}0.71} \\
$\Delta(\text{I}-\text{R})\;\%$ & {\color{tabgreen}\textbf{1.75}} & {\color{tabgreen}\textbf{2.83}} & {\color{tabgreen}\textbf{3.70}} & {\color{tabgreen}\textbf{2.73}} & {\color{tabgreen}\textbf{3.45}} & {\color{tabgreen}\textbf{3.48}} \\
\midrule
$\Delta(\text{I}-\text{M})\;\%$ & {\color{tabgreen}\textbf{2.04}} & {\color{tabgreen}\textbf{2.21}} & {\color{tabgreen}\textbf{1.98}} & {\color{tabgreen}\textbf{1.86}} & {\color{tabgreen}\textbf{2.40}} & {\color{tabgreen}\textbf{2.75}} \\
\bottomrule
\end{tabular}
\label{tab:svhn_noaug}
\end{adjustbox}
\end{table}

\section{Limitations}
The determination of the scale of self-attention ($\sigma$) and the weight normalization step ($\eta$) poses challenges, making the initialization process less reliable. 
Identifying the optimal combination of these two hyperparameters is also challenging. 
Although we attempted SVD decomposition, akin to mimetic initialization, the results were not consistently reliable. 
Instead, we found it more effective to pre-train the initialization of weights for queries and keys, thereby shaping the initial attention map in the form of impulse convolution.

While $\M'_{\I}$ closely resembles the original ViT, minor differences exist. 
Although there are negligible differences in performance, a topic for future work is developing a strategy to initialize attention maps as random impulse convolution filters within the original ViT structure.

\section{Conclusion}
In this paper, we propose a new ViT initialization strategy to address the problem that ViTs are difficult to train on small-scale datasets. 
Our initialization requires no off-line knowledge of pre-trained models on large-scale datasets (mimetic or empirical). 
Our strategy is instead inspired by the inductive bias of convolution, without the need for an architectural modification. 
After a careful study of why random spatial convolution filters work, we opt to initialize self-attention maps as random impulse convolution filters, which incorporate the inductive bias of convolution and preserve the architectural flexibility of transformers. 
Moreover, our approach illuminates a pathway on how to incorporate our knowledge into the construction of self-attention maps.

\clearpage
\setcounter{page}{1}
\maketitlesupplementary

\section{Additional Context on ConvMixer}
\label{apdix:conv}
In the main paper, we provide theoretical insights into the effectiveness of random spatial mixing filters in ConvMixer, highlighting the importance of linear independence and redundancy. 
However, we primarily focus on scenarios where these two conditions are met, introducing the concept of impulse filter initialization. 
In this section, we provide additional experiments to explore cases where linear dependency is not satisfied.
Our main purpose is to demonstrate that, with an adequate amount of redundancy in the number of channels, the expressibility of the spatial mixing convolution layer is intricately linked to the linear independence of the convolution filters.

To reduce the linear independence of the convolution filters, we use Gaussian filters as a soft version of the impulse filter. 
Similar to the impulse filter, a random position in the $f \,{\times}\, f$ filter is chosen, and its value is set to one.
In the impulse filters, all the other elements are zeros, whereas in Gaussian filters, the values around the central one decrease according to the standard deviation $\sigma$. 
Intuitively, when $\sigma$ is very small, the shape of Gaussian closely resembles an impulse filter, and as $\sigma$ increases, the filter gradually becomes flat, resembling averaging filters.

An alternative way to control the linear dependence of the filters is to manually divide the filters into groups and ensure that filters within the same group are identical, such as the concept of ``heads'' in the ViT~\cite{dosovitskiy2020vit}. 
Instead of using $C \,{\times}\, 1 \,{\times}\, f \,{\times}\, f$ parameterized filters, where $C$ represents the number of channels, one can use $H \,{\times}\, 1 \,{\times}\, f \,{\times}\, f$ filters, with $H$ denoting the number of ``heads''.
Each filter here has $\frac{C}{H}$ copies to cover all the channels.

In~\cref{tab:conv_gau}, we show the results of ConvMixer with Gaussian filters. 
The architecture and training settings are consistent with those presented in the main paper. 
The spatial convolution filters are fixed and remain untrained after being initialized as Gaussian filters. 
With a fixed number of channels ($512$), when the number of heads is also $512$, there is no duplication of filters. 
As the standard deviation $\sigma$ decreases from $10$ (wide) to $0.01$ (narrow), the filters become increasingly linearly independent, resulting in a corresponding increase in accuracy.
When the number of heads (different filters) decreases, initially, the performance does not show a significant drop. 
For instance, using $8$ heads performs almost as well as using $512$ heads, suggesting that even with multiple copies of the same filter, good performance can still be achieved within the filter space spanned by $8$ different filters.
However, as this number continues to decrease to $4$ or even $2$, the performance starts to decline as well.

Despite the reduced performance when filters exhibit linear dependence, it is not catastrophically bad, which appears somewhat contradictory to our theoretical expectations. 
However, this observation may be attributed to the relative simplicity of the dataset we used, which may not require complicated spatial mixing.
Additionally, the nonlinearity (GeLU) after this spatial convolution in the spatial convolution layer and the depth of the model (repeating spatial and channel mixing blocks) could potentially alleviate the negative impact of linear dependence in a single spatial mixing layer.

\section{Additional Results of ViT}
\label{apdix:add_result_vit}

In~\cref{tab:svhn_aug} and~\cref{tab:tinyimagenet_aug}, we provide the test accuracy results for random, mimetic~\cite{trockman2023mimetic}, and impulse initialization strategies. 
We keep all other training settings and model architectures the same except for the initialization of the weights for $\Q$ and $\K$. 
We observe that the order of the three different initialization strategies aligns with our findings presented in the main paper.

\begin{table}[t]
\caption[]{Classification accuracy(\%) on augmented CIFAR-10~\cite{krizhevsky2009learning} of Gaussian filters ConvMixer with different standard deviation $\sigma$ and the number of different filters,~\ie, ``heads''.}
\centering
\begin{adjustbox}{width=0.8\linewidth}
\begin{tabular}{lcccccccc}
\toprule

\multirow{2}{*}{$\sigma$} & \multicolumn{5}{c}{heads}              \\ \cline{2-6} 
                          & 512   & 64    & 8     & 4     & 2     \\
\midrule
10                        & 83.35 & 83.4  & 83.45 & 83.05 & 82.39 \\
1                         & 88.49 & 88.25 & 88.07 & 87.93 & 87.52 \\
0.5                       & 89.22 & 89.07 & 89.32 & 88.60  & 88.90  \\
0.3                       & 90.57 & 90.74 & 90.58 & 89.85 & 88.06 \\
0.1                       & 91.48 & 91.7  & 91.15 & 90.45 & 89.07 \\
0.01                      & 91.60  & 91.73 & 91.82 & 90.14 & 89.98
\\
\bottomrule
\end{tabular}
\label{tab:conv_gau}
\end{adjustbox}
\end{table}

\begin{table}[t]
\caption[]{Classification accuracy(\%) on augmented SVHN~\cite{yuval2011reading} of $\M'_{\I}$ in~\cref{equ:alpha_m1} with different initialization strategies and $\alpha$s. 
When calculating difference $\Delta$s, ``R'' stands for random, ``M'' denotes mimetic, and ``I'' represents our impulse initialization.
{\color{tabgreen}\textbf{Green}} numbers indicate the percentage of accuracy increase.}
\centering
\begin{adjustbox}{width=\linewidth}
\begin{tabular}{lcccccccc}
\toprule
\thead{\normalsize $\alpha$} & \thead{\normalsize 0.0} & \thead{\normalsize 0.1} & \thead{\normalsize 0.2} & \thead{\normalsize 0.3} & \thead{\normalsize 0.4} & \thead{\normalsize 0.5}\\
\midrule
Random & 95.43 & 95.66 & 95.35 & 94.82 & 94.81 & 95.24\\
Mimetic~\cite{trockman2023mimetic} & 95.44 & 96.15 & 96.02 & 95.75 & 95.73 & 95.71\\
Impulse (Ours) & 95.65 & 96.21 & 96.19 & 96.25 & 96.35 & 96.15\\
\midrule
$\Delta(\text{M}-\text{R})\;\%$ & {\color{tabgreen}0.01} & {\color{tabgreen}0.51} & {\color{tabgreen}0.70} & {\color{tabgreen}0.98} & {\color{tabgreen}0.97} & {\color{tabgreen}0.49} \\
$\Delta(\text{I}-\text{R})\;\%$ & {\color{tabgreen}\textbf{0.23}} & {\color{tabgreen}\textbf{0.57}} & {\color{tabgreen}\textbf{0.88}} & {\color{tabgreen}\textbf{1.51}} & {\color{tabgreen}\textbf{1.62}} & {\color{tabgreen}\textbf{0.96}} \\
\midrule
$\Delta(\text{I}-\text{M})\;\%$ & {\color{tabgreen}\textbf{0.22}} & {\color{tabgreen}\textbf{0.06}} & {\color{tabgreen}\textbf{0.18}} & {\color{tabgreen}\textbf{0.52}} & {\color{tabgreen}\textbf{0.64}} & {\color{tabgreen}\textbf{0.46}} \\
\bottomrule
\end{tabular}
\label{tab:svhn_aug}
\end{adjustbox}
\end{table}

\begin{table}[t]
\caption[]{Classification accuracy(\%) on augmented Tiny-ImageNet~\cite{le2015tiny} of $\M'_{\I}$ in~\cref{equ:alpha_m1} with different initialization strategies and $\alpha$s. 
When calculating the difference $\Delta$, ``R'' stands for random, ``M'' denotes mimetic, and ``I'' represents our impulse initialization.
{\color{tabgreen}\textbf{Green}} numbers indicate the percentage of accuracy increase.}
\centering
\begin{adjustbox}{width=\linewidth}
\begin{tabular}{lcccccccc}
\toprule
\thead{\normalsize $\alpha$} & \thead{\normalsize 0.0} & \thead{\normalsize 0.1} & \thead{\normalsize 0.2} & \thead{\normalsize 0.3} & \thead{\normalsize 0.4} & \thead{\normalsize 0.5}\\
\midrule
Random & 47.71 & 50.39 & 49.50 & 50.03 & 47.65 & 46.74\\
Mimetic~\cite{trockman2023mimetic} &  48.74 & 50.65 & 50.70 & 50.41 & 49.80 & 48.90\\
Impulse (Ours) &  48.76 & 52.84 & 52.43 & 51.55 & 51.93 & 50.89\\
\midrule
$\Delta(\text{M}-\text{R})\;\%$ & {\color{tabgreen}2.16} & {\color{tabgreen}0.52} & {\color{tabgreen}2.42} & {\color{tabgreen}0.76} & {\color{tabgreen}4.51} & {\color{tabgreen}4.62} \\
$\Delta(\text{I}-\text{R})\;\%$ & {\color{tabgreen}\textbf{2.20}} & {\color{tabgreen}\textbf{4.86}} & {\color{tabgreen}\textbf{5.92}} & {\color{tabgreen}\textbf{3.04}} & {\color{tabgreen}\textbf{8.98}} & {\color{tabgreen}\textbf{8.88}} \\
\midrule
$\Delta(\text{I}-\text{M})\;\%$ & {\color{tabgreen}\textbf{0.04}} & {\color{tabgreen}\textbf{4.32}} & {\color{tabgreen}\textbf{3.41}} & {\color{tabgreen}\textbf{2.26}} & {\color{tabgreen}\textbf{4.28}} & {\color{tabgreen}\textbf{4.07}} \\
\bottomrule
\end{tabular}
\label{tab:tinyimagenet_aug}
\end{adjustbox}
\end{table}

\section{Visualization of different initialization strategies}
\label{apdix:vil_vit}

\begin{figure*}[ht]
%\captionsetup{size=small}
\captionsetup[subfigure]{labelformat=empty}
\centering

% \begin{subfigure}{0.2\textwidth}\hspace{-2mm}
% \centering
% \rotatebox[origin=a]{90}{\centering$\Q\K$}
% \end{subfigure}
\begin{subfigure}{0.19\textwidth}
\includegraphics[width=\linewidth]{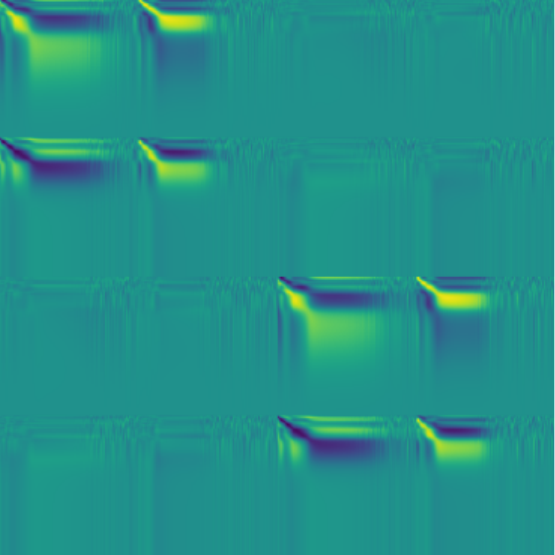}
%\vspace{-1.5\baselineskip}
\end{subfigure} 
\begin{subfigure}{0.19\textwidth}
\includegraphics[height=\linewidth]{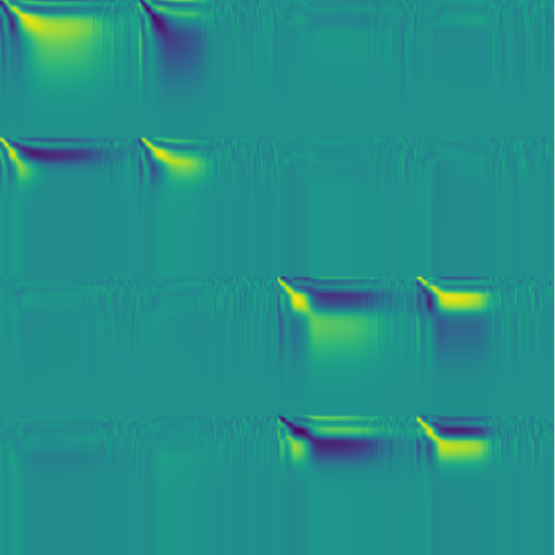}
%\vspace{-1.5\baselineskip}
\end{subfigure}
\begin{subfigure}{0.19\textwidth}
\includegraphics[height=\linewidth]{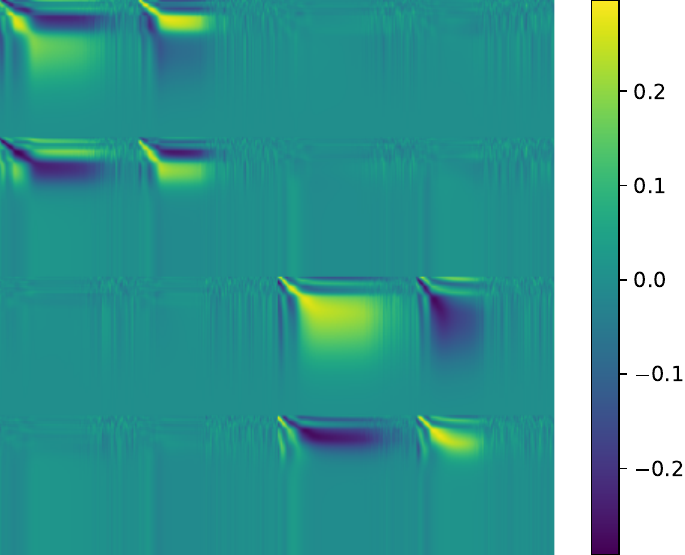}
%\vspace{-1.5\baselineskip}
\end{subfigure}

% \begin{subfigure}{0.19\textwidth}\hspace{-1mm}
% \rotatebox[origin=c]{90}{\centering\textbf{LogF}}
% \end{subfigure}
\begin{subfigure}{0.19\textwidth}
\includegraphics[height=\linewidth]{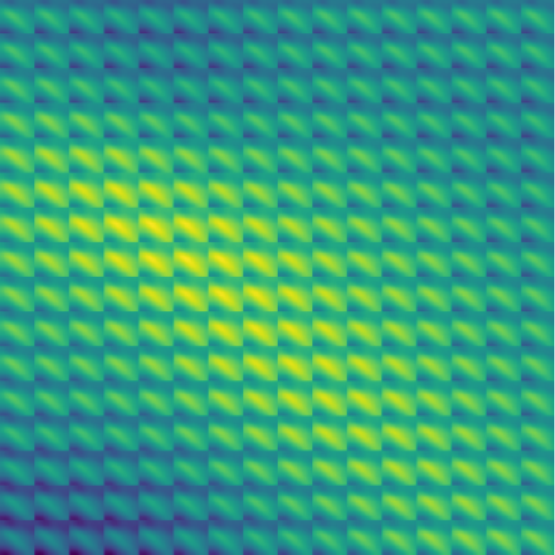} 
%\vspace{-1.5\baselineskip}
\end{subfigure} 
\begin{subfigure}{0.19\textwidth}
\includegraphics[height=\linewidth]{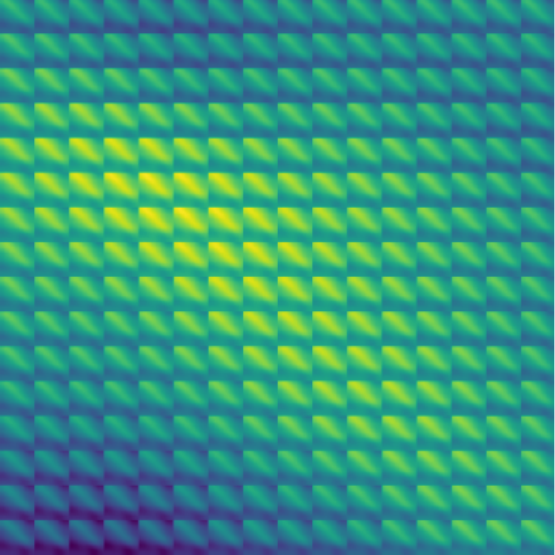} 
%\vspace{-1.5\baselineskip}
\end{subfigure} 
\begin{subfigure}{0.19\textwidth}
\includegraphics[height=\linewidth]{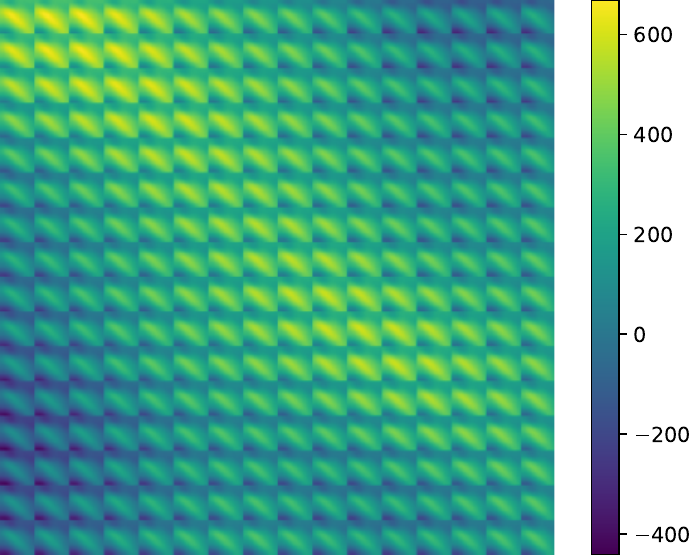} 
%\vspace{-1.5\baselineskip}
\end{subfigure}

% \begin{subfigure}{0.015\textwidth}\hspace{-1mm}
% \rotatebox[origin=c]{90}{\textbf{LogF}}
% \end{subfigure}
\begin{subfigure}{0.19\textwidth}
\includegraphics[height=\linewidth]{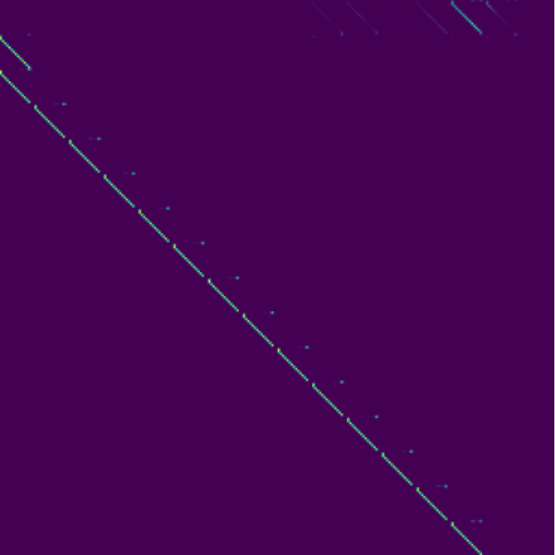} 
%\vspace{-1.0\baselineskip}
\caption{\textbf{Layer 1}}
%\vspace{-0.5\baselineskip}
\end{subfigure} 
\begin{subfigure}{0.19\textwidth}
\includegraphics[height=\linewidth]{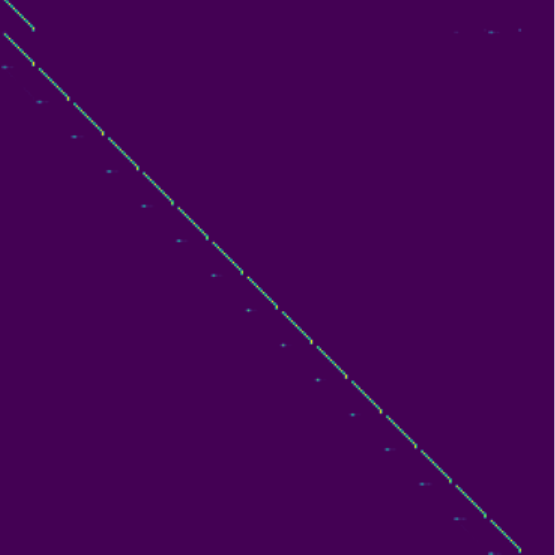} 
%\vspace{-1.0\baselineskip}
\caption{\textbf{Layer 3}}
%\vspace{-0.5\baselineskip}
\end{subfigure} 
\begin{subfigure}{0.19\textwidth}
\includegraphics[height=\linewidth]{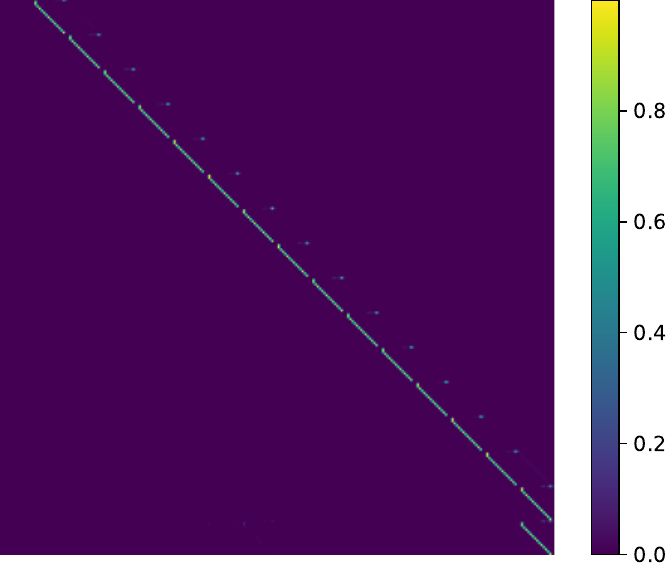} 
%\vspace{-1.0\baselineskip}
\caption{\textbf{Layer 6}}
%\vspace{-0.5\baselineskip}
\end{subfigure}

% \begin{subfigure}{0.015\textwidth}\hspace{-2mm}
% \rotatebox[origin=c]{90}{$\Q\K$}
% \end{subfigure}
\begin{subfigure}{0.19\textwidth}
\includegraphics[width=\linewidth]{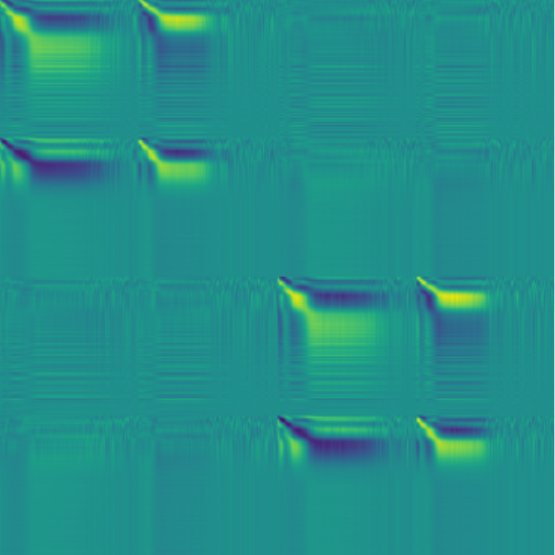}
%\vspace{-1.5\baselineskip}
\end{subfigure} 
\begin{subfigure}{0.19\textwidth}
\includegraphics[height=\linewidth]{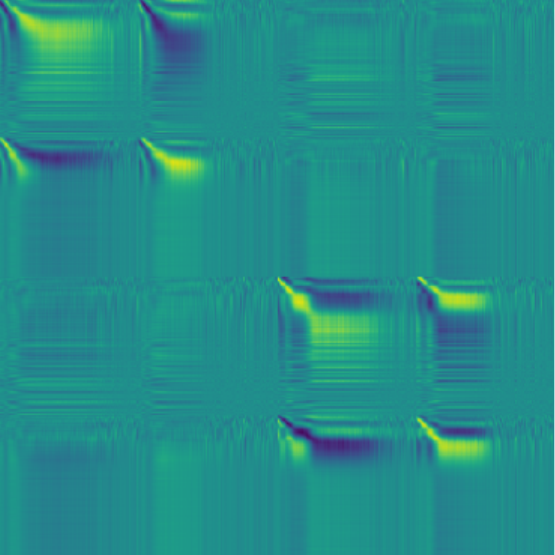}
%\vspace{-1.5\baselineskip}
\end{subfigure}
\begin{subfigure}{0.19\textwidth}
\includegraphics[height=\linewidth]{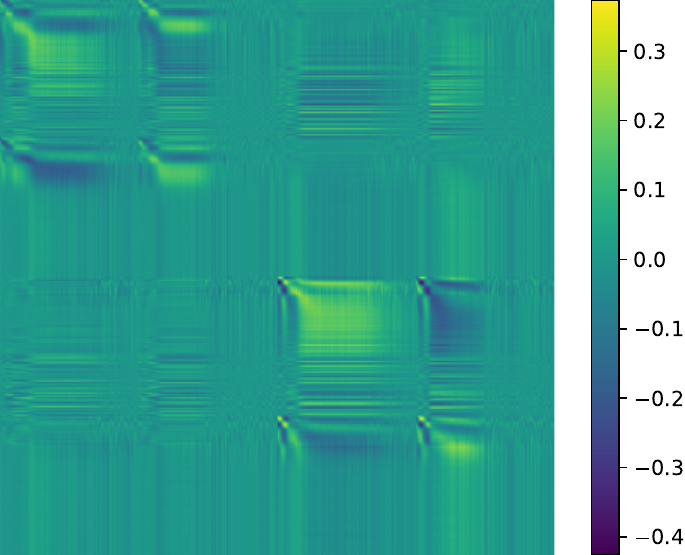}
%\vspace{-1.5\baselineskip}
\end{subfigure}

% \begin{subfigure}{0.015\textwidth}\hspace{-1mm}
% \rotatebox[origin=c]{90}{\textbf{LogF}}
% \end{subfigure}
\begin{subfigure}{0.19\textwidth}
\includegraphics[height=\linewidth]{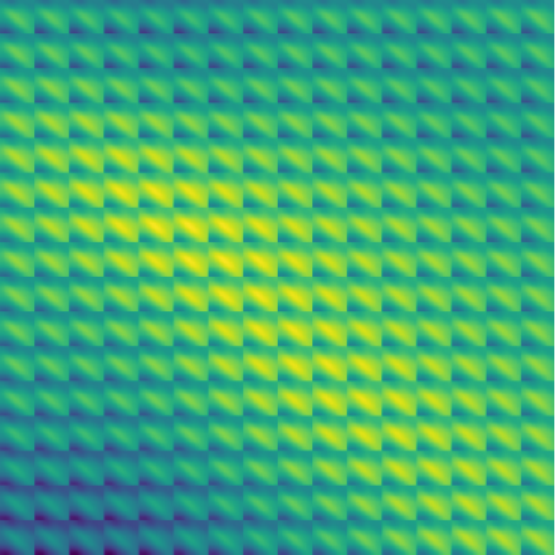} 
%\vspace{-1.5\baselineskip}
\end{subfigure} 
\begin{subfigure}{0.19\textwidth}
\includegraphics[height=\linewidth]{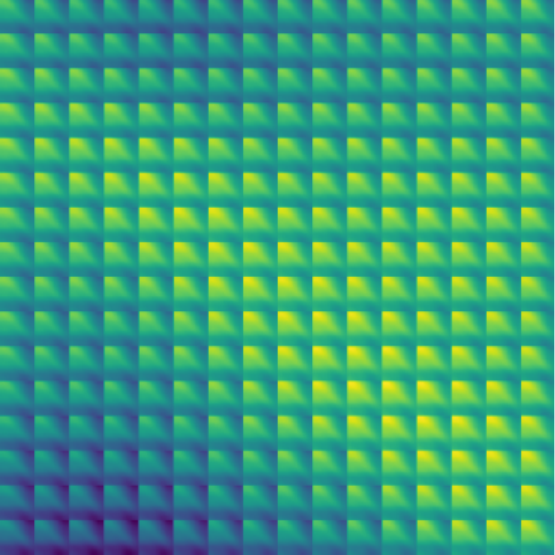} 
%\vspace{-1.5\baselineskip}
\end{subfigure} 
\begin{subfigure}{0.19\textwidth}
\includegraphics[height=\linewidth]{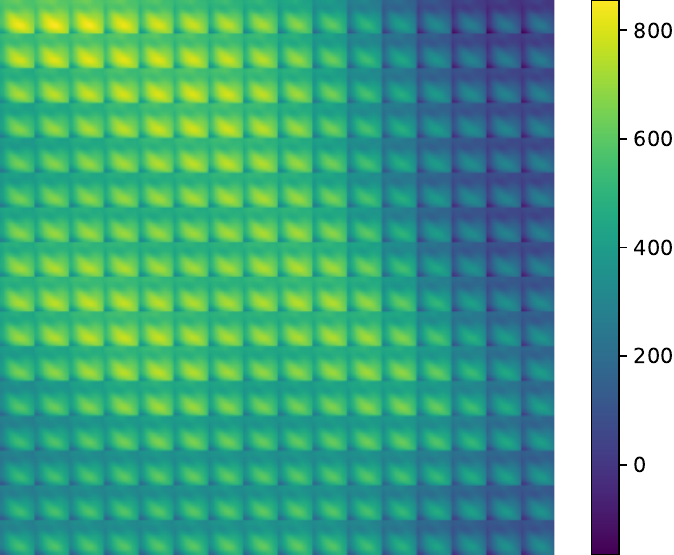} 
%\vspace{-1.5\baselineskip}
\end{subfigure}

% \begin{subfigure}{0.015\textwidth}\hspace{-1mm}
% \rotatebox[origin=c]{90}{\textbf{LogF}}
% \end{subfigure}
\begin{subfigure}{0.19\textwidth}
\includegraphics[height=\linewidth]{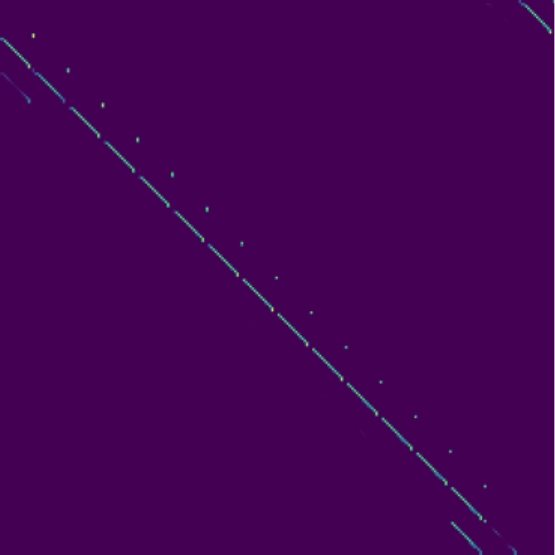} 
%\vspace{-1.5\baselineskip}
\end{subfigure} 
\begin{subfigure}{0.19\textwidth}
\includegraphics[height=\linewidth]{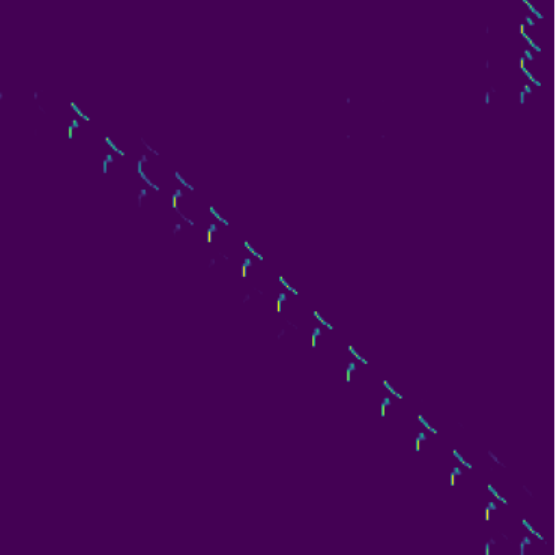} 
%\vspace{-1.5\baselineskip}
\end{subfigure} 
\begin{subfigure}{0.19\textwidth}
\includegraphics[height=\linewidth]{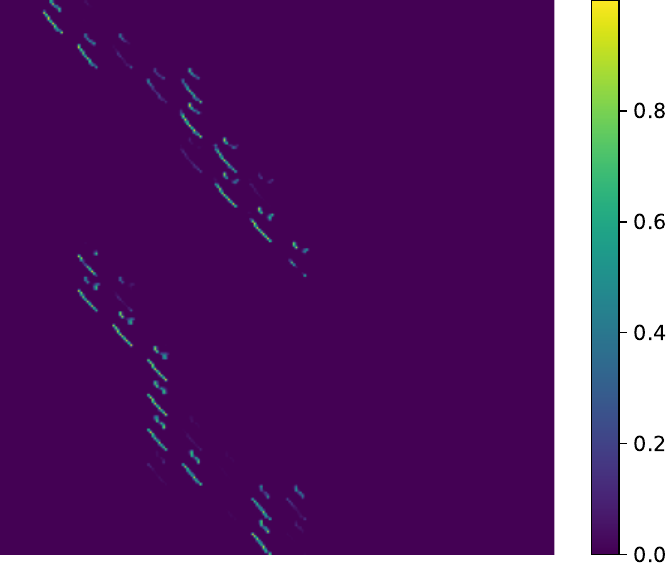} 
%\vspace{-1.5\baselineskip}
\end{subfigure}

\caption{Visualization of $\Q$ and $\K$ with impulse initialization. 
The upper section of the figure is initialization (before training), while the lower section is trained. 
In each section, the first row is $\Q\K$, the second row is $\P\Q\K\P^{T}$, and the third row is $\mbox{SoftMax}(\P\Q\K\P^{T})$. 
Columns are laid out as layers 1, 3, and 6 (depth is 6). 
Because the 8 heads in each layer have similar patterns, we only choose head 3 for illustration for all the figures.}
\label{fig:impulse_init}
\end{figure*}

\begin{figure*}[ht]
%\captionsetup{size=small}
\captionsetup[subfigure]{labelformat=empty}
\centering

% \begin{subfigure}{0.2\textwidth}\hspace{-2mm}
% \centering
% \rotatebox[origin=a]{90}{\centering$\Q\K$}
% \end{subfigure}
\begin{subfigure}{0.19\textwidth}
\includegraphics[width=\linewidth]{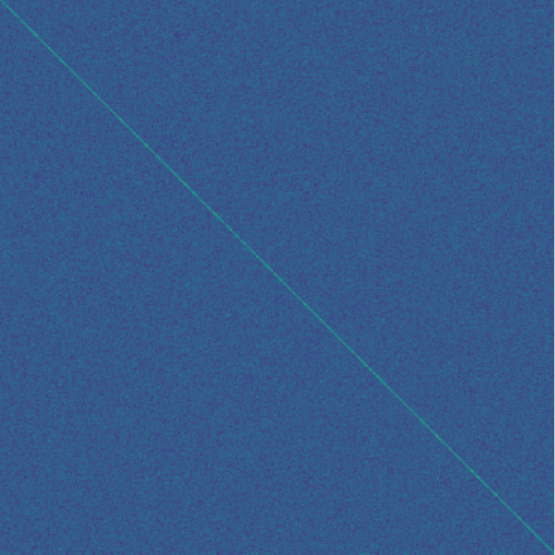}
%\vspace{-1.5\baselineskip}
\end{subfigure} 
\begin{subfigure}{0.19\textwidth}
\includegraphics[height=\linewidth]{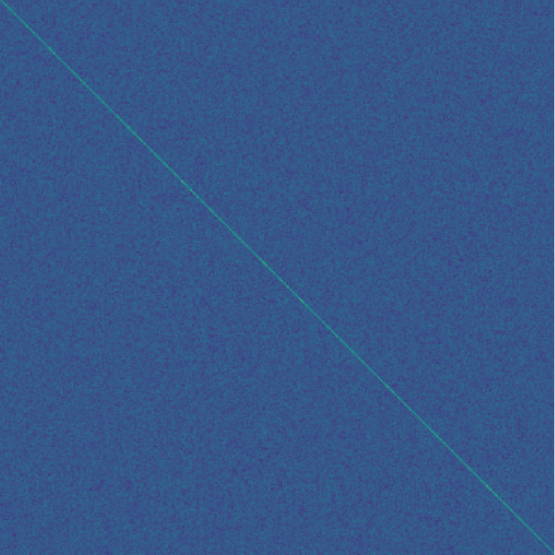}
%\vspace{-1.5\baselineskip}
\end{subfigure}
\begin{subfigure}{0.19\textwidth}
\includegraphics[height=\linewidth]{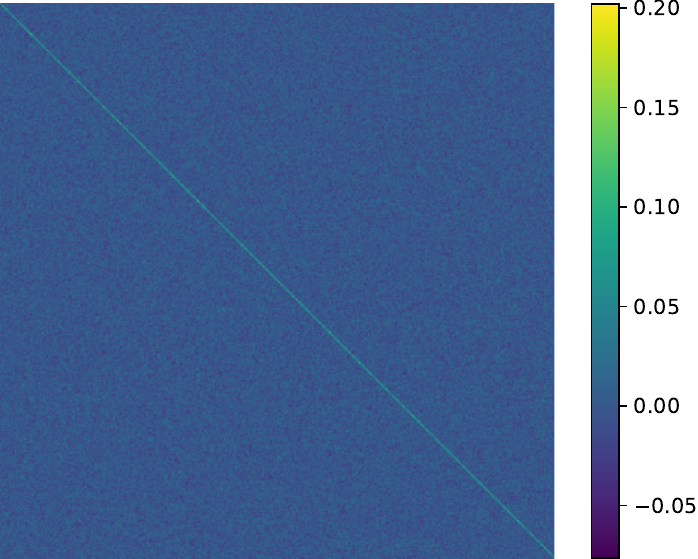}
%\vspace{-1.5\baselineskip}
\end{subfigure}

% \begin{subfigure}{0.19\textwidth}\hspace{-1mm}
% \rotatebox[origin=c]{90}{\centering\textbf{LogF}}
% \end{subfigure}
\begin{subfigure}{0.19\textwidth}
\includegraphics[height=\linewidth]{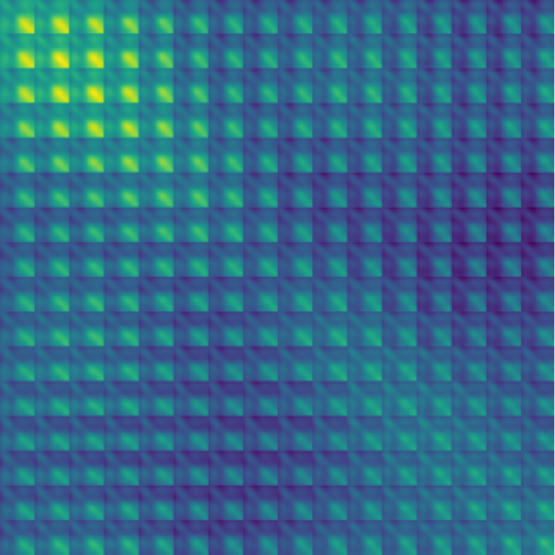} 
%\vspace{-1.5\baselineskip}
\end{subfigure} 
\begin{subfigure}{0.19\textwidth}
\includegraphics[height=\linewidth]{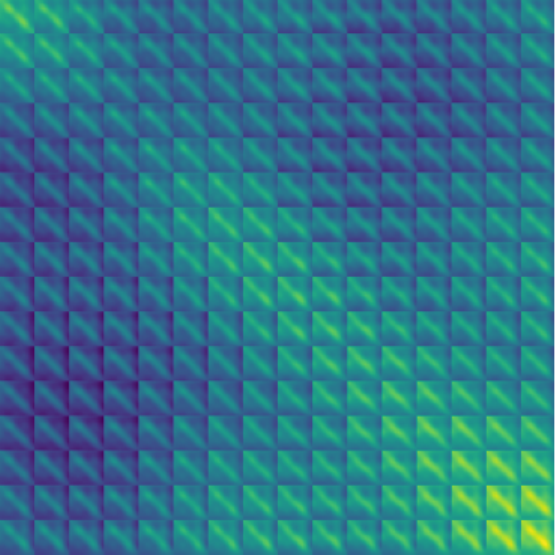} 
%\vspace{-1.5\baselineskip}
\end{subfigure} 
\begin{subfigure}{0.19\textwidth}
\includegraphics[height=\linewidth]{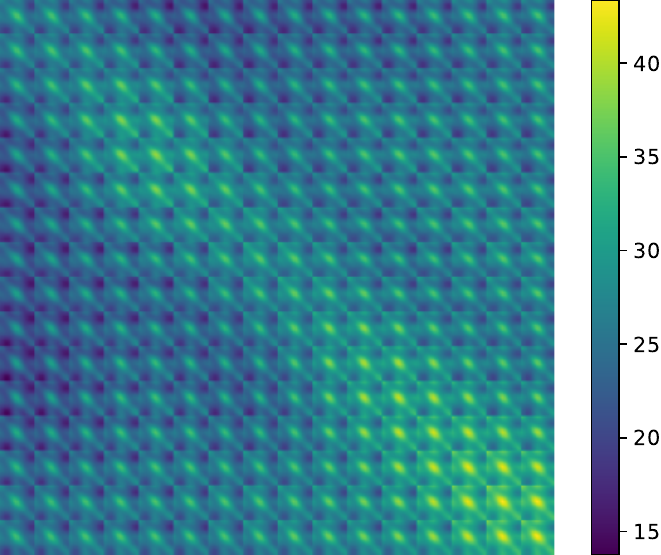} 
%\vspace{-1.5\baselineskip}
\end{subfigure}

% \begin{subfigure}{0.015\textwidth}\hspace{-1mm}
% \rotatebox[origin=c]{90}{\textbf{LogF}}
% \end{subfigure}
\begin{subfigure}{0.19\textwidth}
\includegraphics[height=\linewidth]{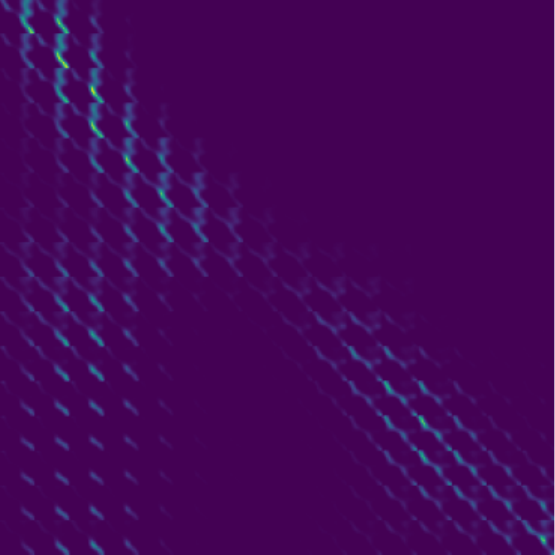} 
%\vspace{-1.0\baselineskip}
\caption{\textbf{Layer 1}}
%\vspace{-0.5\baselineskip}
\end{subfigure} 
\begin{subfigure}{0.19\textwidth}
\includegraphics[height=\linewidth]{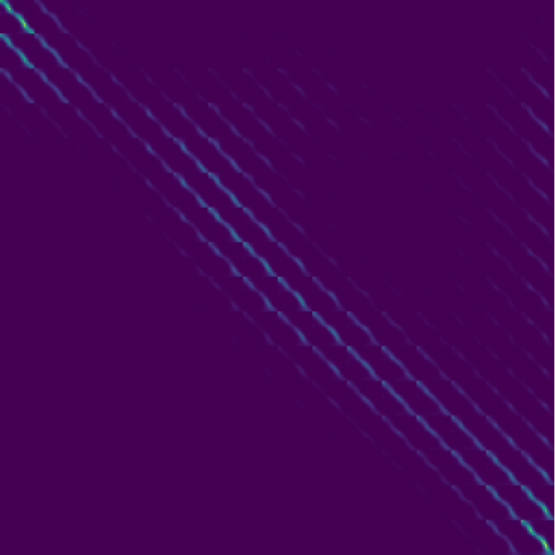} 
%\vspace{-1.0\baselineskip}
\caption{\textbf{Layer 3}}
%\vspace{-0.5\baselineskip}
\end{subfigure} 
\begin{subfigure}{0.19\textwidth}
\includegraphics[height=\linewidth]{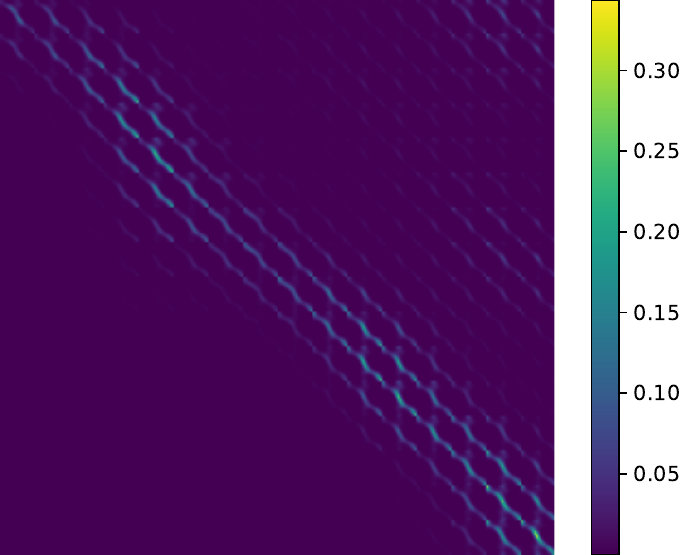} 
%\vspace{-1.0\baselineskip}
\caption{\textbf{Layer 6}}
%\vspace{-0.5\baselineskip}
\end{subfigure}

% \begin{subfigure}{0.015\textwidth}\hspace{-2mm}
% \rotatebox[origin=c]{90}{$\Q\K$}
% \end{subfigure}
\begin{subfigure}{0.19\textwidth}
\includegraphics[width=\linewidth]{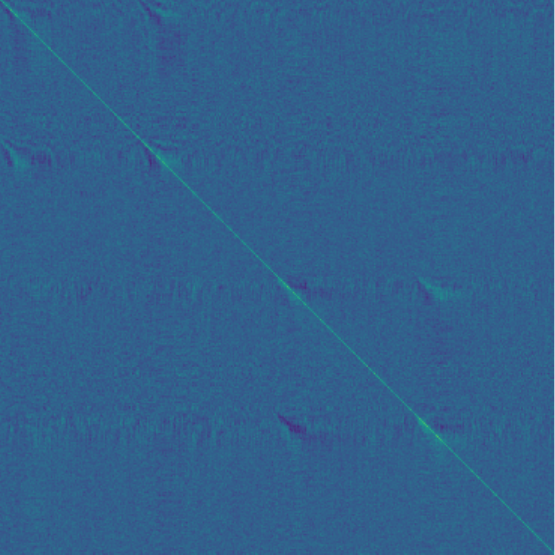}
%\vspace{-1.5\baselineskip}
\end{subfigure} 
\begin{subfigure}{0.19\textwidth}
\includegraphics[height=\linewidth]{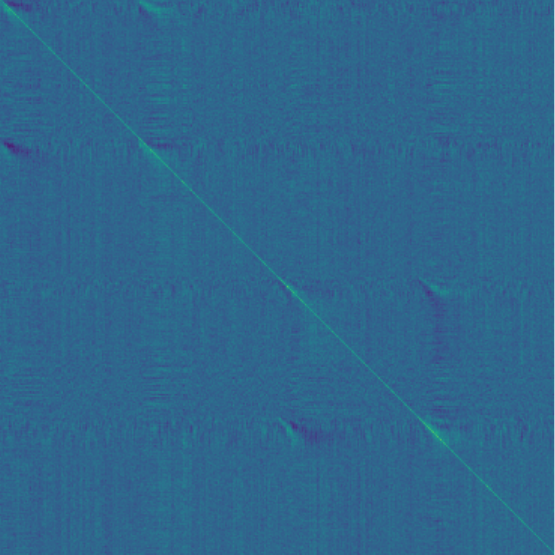}
%\vspace{-1.5\baselineskip}
\end{subfigure}
\begin{subfigure}{0.19\textwidth}
\includegraphics[height=\linewidth]{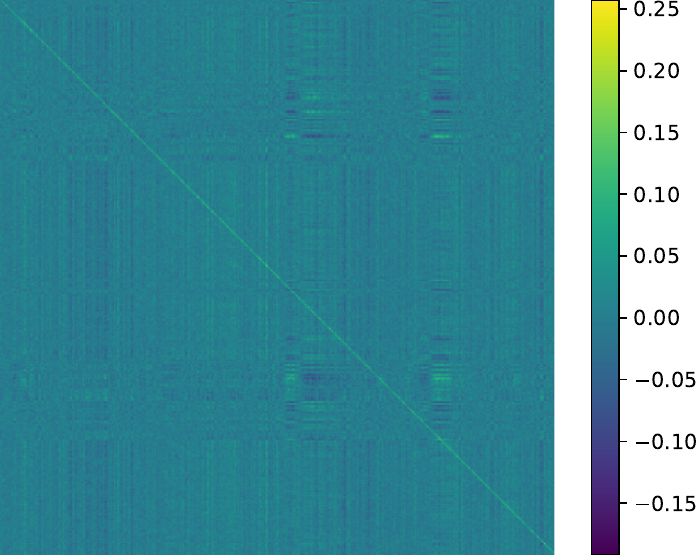}
%\vspace{-1.5\baselineskip}
\end{subfigure}

% \begin{subfigure}{0.015\textwidth}\hspace{-1mm}
% \rotatebox[origin=c]{90}{\textbf{LogF}}
% \end{subfigure}
\begin{subfigure}{0.19\textwidth}
\includegraphics[height=\linewidth]{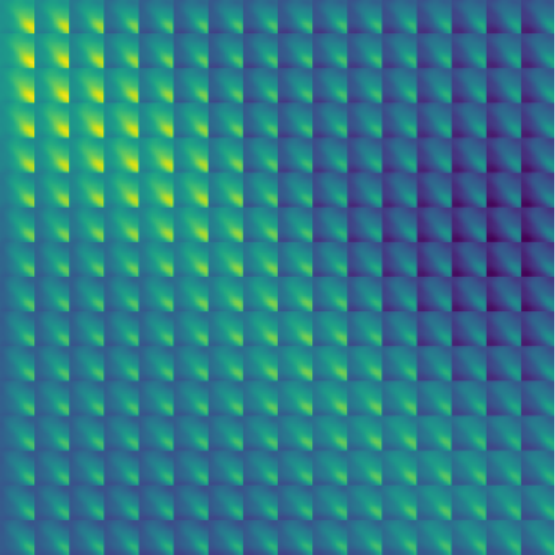} 
%\vspace{-1.5\baselineskip}
\end{subfigure} 
\begin{subfigure}{0.19\textwidth}
\includegraphics[height=\linewidth]{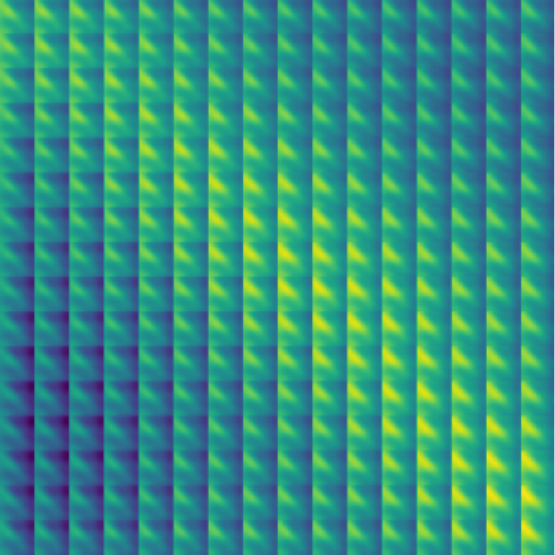} 
%\vspace{-1.5\baselineskip}
\end{subfigure} 
\begin{subfigure}{0.19\textwidth}
\includegraphics[height=\linewidth]{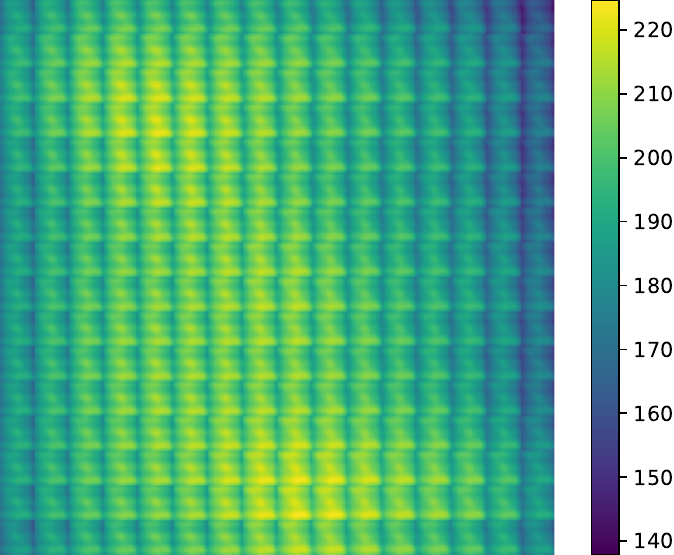} 
%\vspace{-1.5\baselineskip}
\end{subfigure}

% \begin{subfigure}{0.015\textwidth}\hspace{-1mm}
% \rotatebox[origin=c]{90}{\textbf{LogF}}
% \end{subfigure}
\begin{subfigure}{0.19\textwidth}
\includegraphics[height=\linewidth]{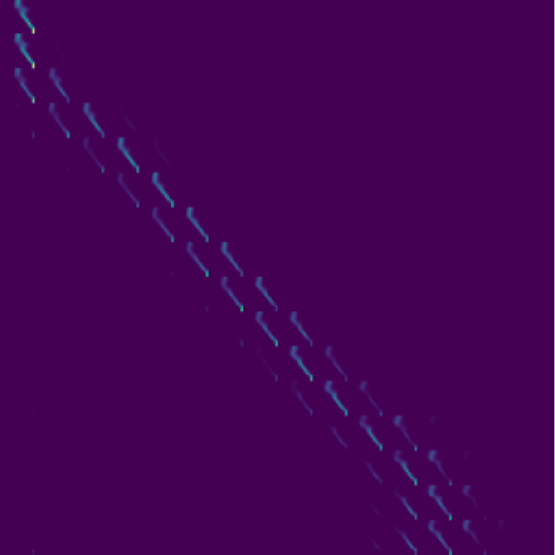} 
%\vspace{-1.5\baselineskip}
\end{subfigure} 
\begin{subfigure}{0.19\textwidth}
\includegraphics[height=\linewidth]{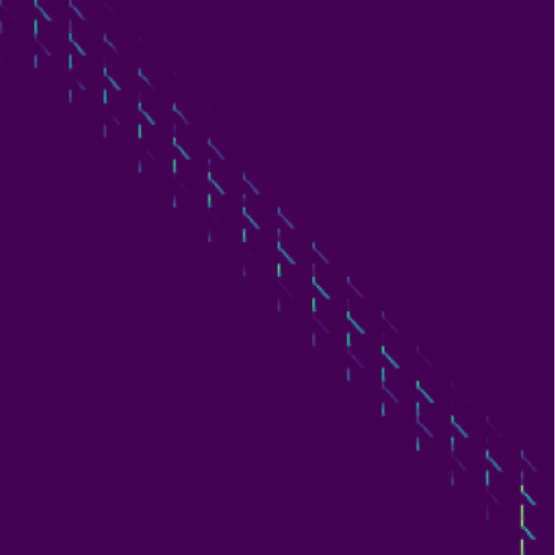} 
%\vspace{-1.5\baselineskip}
\end{subfigure} 
\begin{subfigure}{0.19\textwidth}
\includegraphics[height=\linewidth]{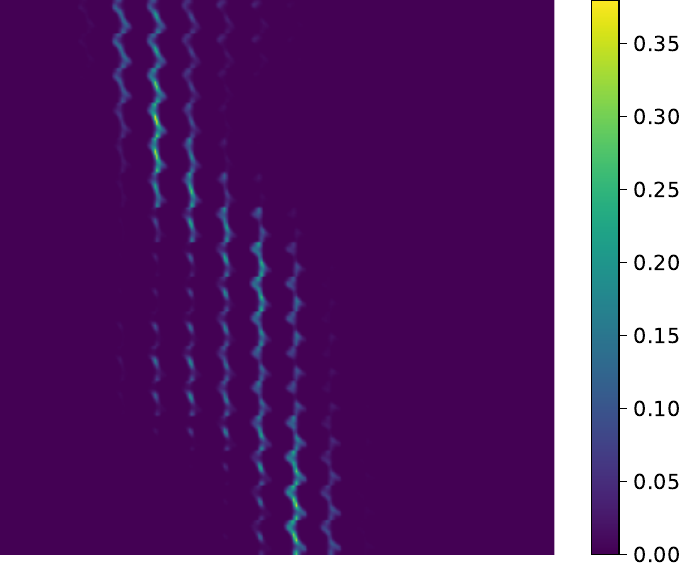} 
%\vspace{-1.5\baselineskip}
\end{subfigure}

\caption{Visualization of $\Q$ and $\K$ with mimetic initialization. 
The upper section of the figure is initialization (before training), while the lower section is trained. 
In each section, the first row is $\Q\K$, the second row is $\P\Q\K\P^{T}$, and the third row is $\mbox{SoftMax}(\P\Q\K\P^{T})$. 
Columns are laid out as layers 1, 3, and 6 (depth is 6). 
Because the 8 heads in each layer have similar patterns, we only choose head 3 for illustration for all the figures.}
\label{fig:mimetic_init}
\end{figure*}

\begin{figure*}[ht]
%\captionsetup{size=small}
\captionsetup[subfigure]{labelformat=empty}
\centering

% \begin{subfigure}{0.19\textwidth}\hspace{-2mm}
% \centering
% \rotatebox[origin=a]{90}{\centering$\Q\K$}
% \end{subfigure}
\begin{subfigure}{0.19\textwidth}
\includegraphics[width=\linewidth]{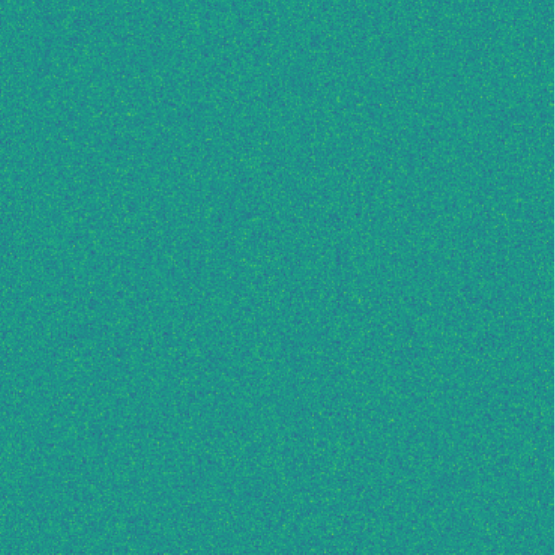}
%\vspace{-1.5\baselineskip}
\end{subfigure} 
\begin{subfigure}{0.19\textwidth}
\includegraphics[height=\linewidth]{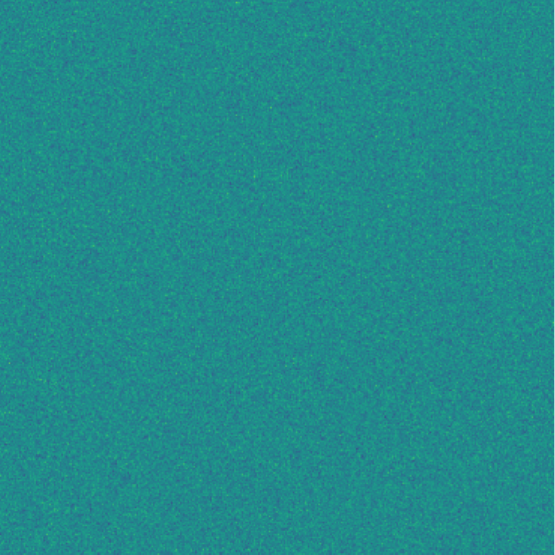}
%\vspace{-1.5\baselineskip}
\end{subfigure}
\begin{subfigure}{0.19\textwidth}
\includegraphics[height=\linewidth]{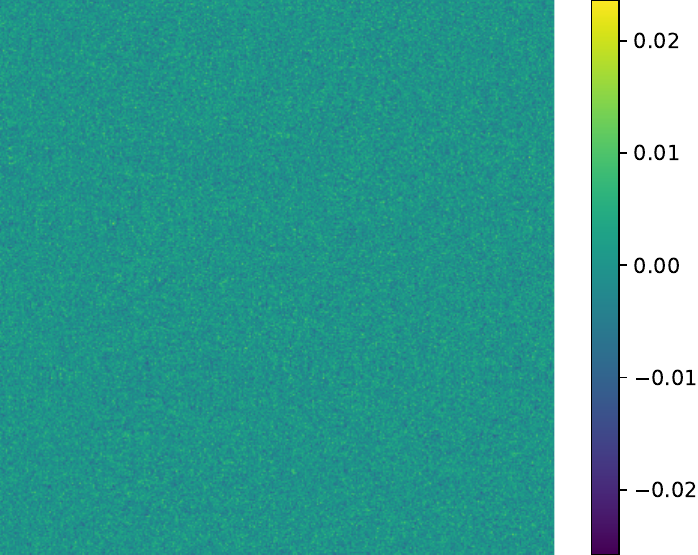}
%\vspace{-1.5\baselineskip}
\end{subfigure}

% \begin{subfigure}{0.19\textwidth}\hspace{-1mm}
% \rotatebox[origin=c]{90}{\centering\textbf{LogF}}
% \end{subfigure}
\begin{subfigure}{0.19\textwidth}
\includegraphics[height=\linewidth]{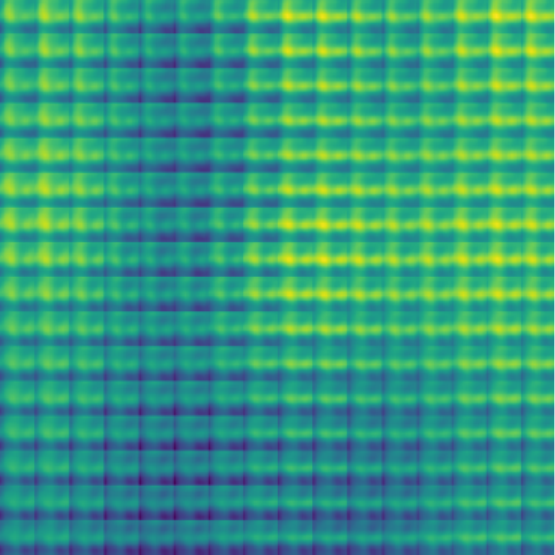} 
%\vspace{-1.5\baselineskip}
\end{subfigure} 
\begin{subfigure}{0.19\textwidth}
\includegraphics[height=\linewidth]{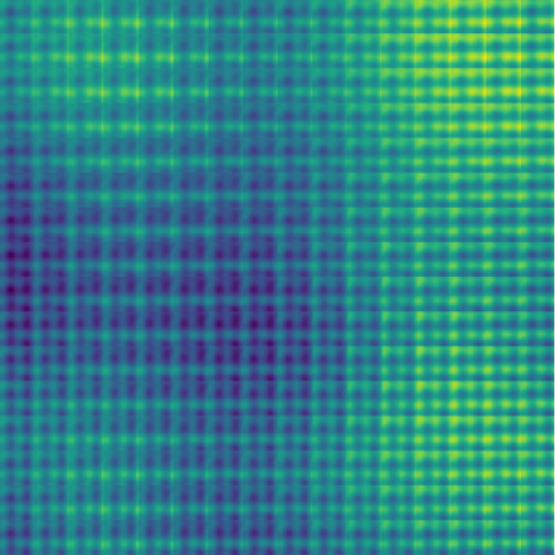} 
%\vspace{-1.5\baselineskip}
\end{subfigure} 
\begin{subfigure}{0.19\textwidth}
\includegraphics[height=\linewidth]{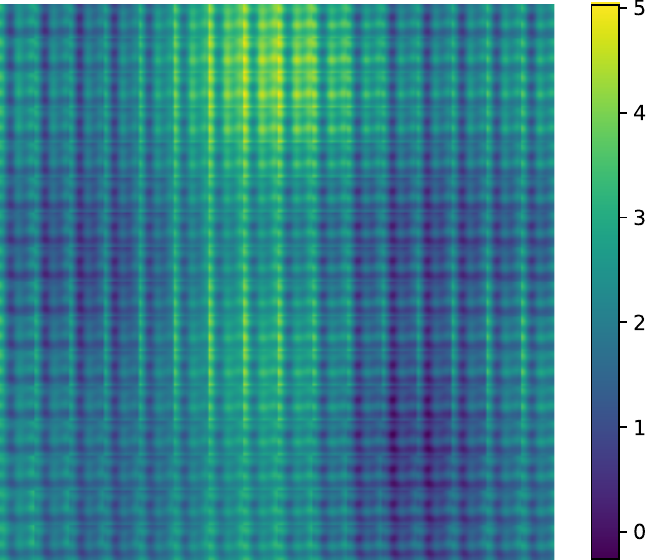} 
%\vspace{-1.5\baselineskip}
\end{subfigure}

% \begin{subfigure}{0.015\textwidth}\hspace{-1mm}
% \rotatebox[origin=c]{90}{\textbf{LogF}}
% \end{subfigure}
\begin{subfigure}{0.19\textwidth}
\includegraphics[height=\linewidth]{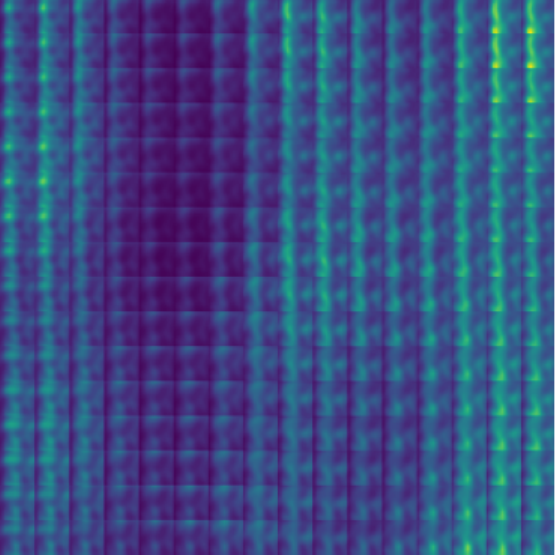} 
%\vspace{-1.0\baselineskip}
\caption{\textbf{Layer 1}}
%\vspace{-0.5\baselineskip}
\end{subfigure} 
\begin{subfigure}{0.19\textwidth}
\includegraphics[height=\linewidth]{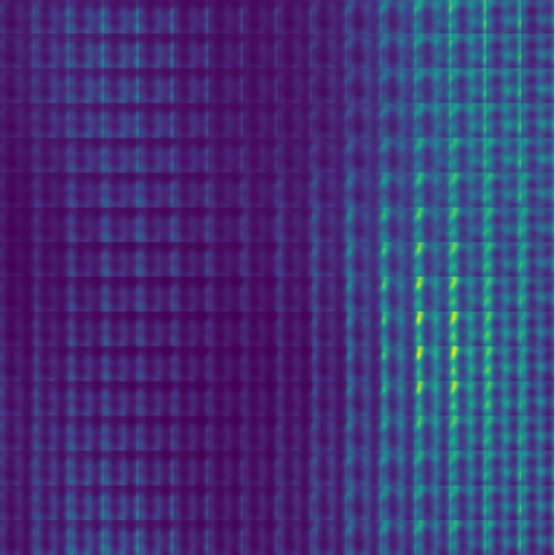} 
%\vspace{-1.0\baselineskip}
\caption{\textbf{Layer 3}}
%\vspace{-0.5\baselineskip}
\end{subfigure} 
\begin{subfigure}{0.19\textwidth}
\includegraphics[height=\linewidth]{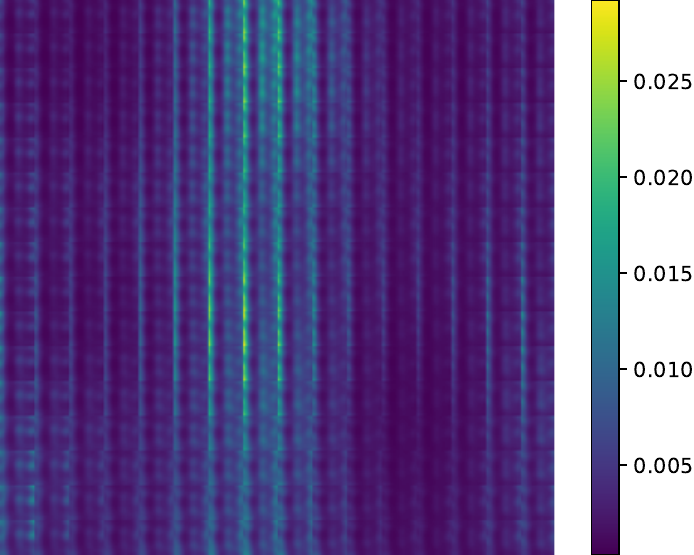} 
%\vspace{-1.0\baselineskip}
\caption{\textbf{Layer 6}}
%\vspace{-0.5\baselineskip}
\end{subfigure}

% \begin{subfigure}{0.015\textwidth}\hspace{-2mm}
% \rotatebox[origin=c]{90}{$\Q\K$}
% \end{subfigure}
\begin{subfigure}{0.19\textwidth}
\includegraphics[width=\linewidth]{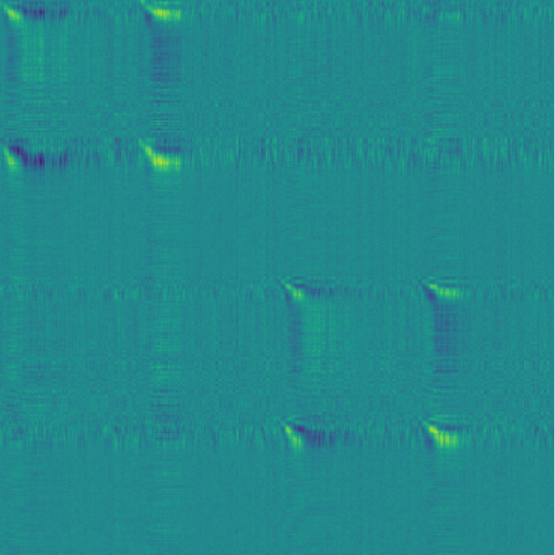}
%\vspace{-1.5\baselineskip}
\end{subfigure} 
\begin{subfigure}{0.19\textwidth}
\includegraphics[height=\linewidth]{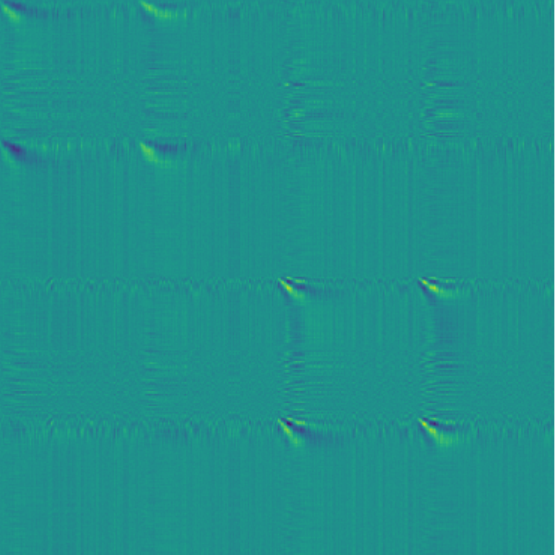}
%\vspace{-1.5\baselineskip}
\end{subfigure}
\begin{subfigure}{0.19\textwidth}
\includegraphics[height=\linewidth]{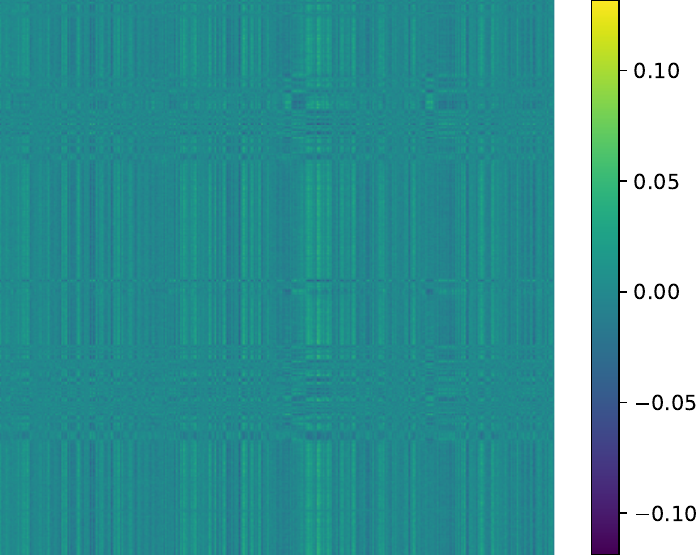}
%\vspace{-1.5\baselineskip}
\end{subfigure}

% \begin{subfigure}{0.015\textwidth}\hspace{-1mm}
% \rotatebox[origin=c]{90}{\textbf{LogF}}
% \end{subfigure}
\begin{subfigure}{0.19\textwidth}
\includegraphics[height=\linewidth]{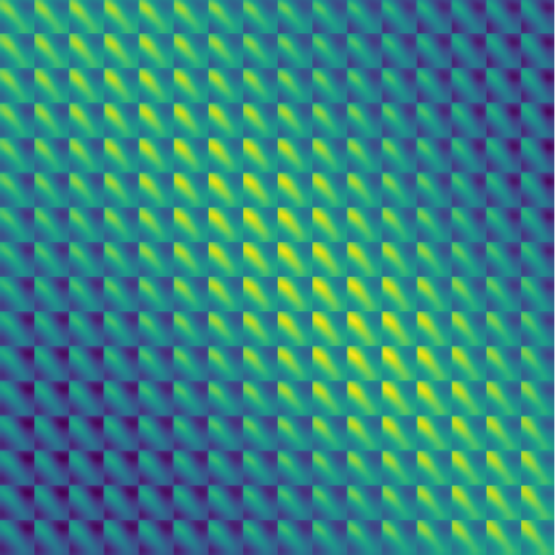} 
%\vspace{-1.5\baselineskip}
\end{subfigure} 
\begin{subfigure}{0.19\textwidth}
\includegraphics[height=\linewidth]{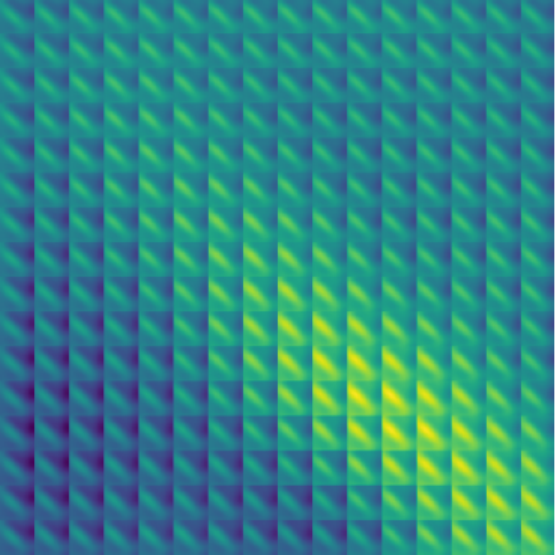} 
%\vspace{-1.5\baselineskip}
\end{subfigure} 
\begin{subfigure}{0.19\textwidth}
\includegraphics[height=\linewidth]{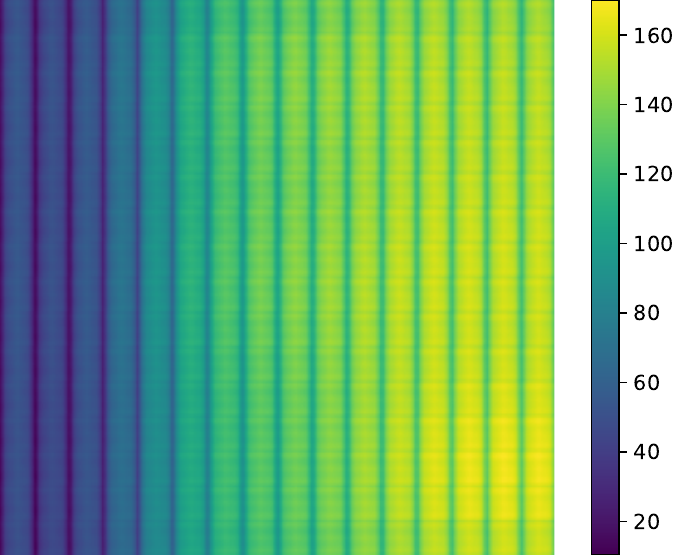} 
%\vspace{-1.5\baselineskip}
\end{subfigure}

% \begin{subfigure}{0.015\textwidth}\hspace{-1mm}
% \rotatebox[origin=c]{90}{\textbf{LogF}}
% \end{subfigure}
\begin{subfigure}{0.19\textwidth}
\includegraphics[height=\linewidth]{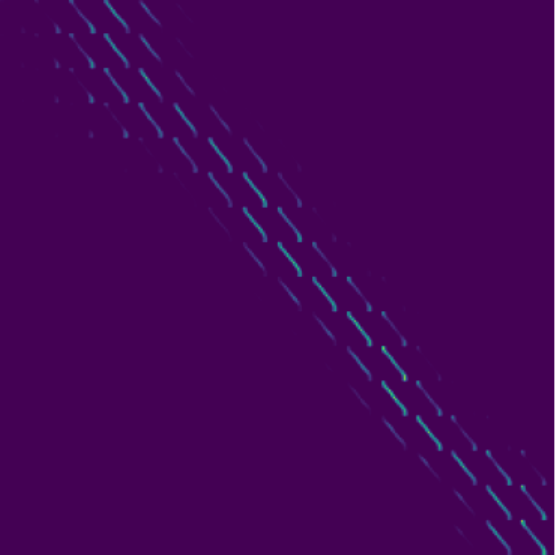} 
%\vspace{-1.5\baselineskip}
\end{subfigure} 
\begin{subfigure}{0.19\textwidth}
\includegraphics[height=\linewidth]{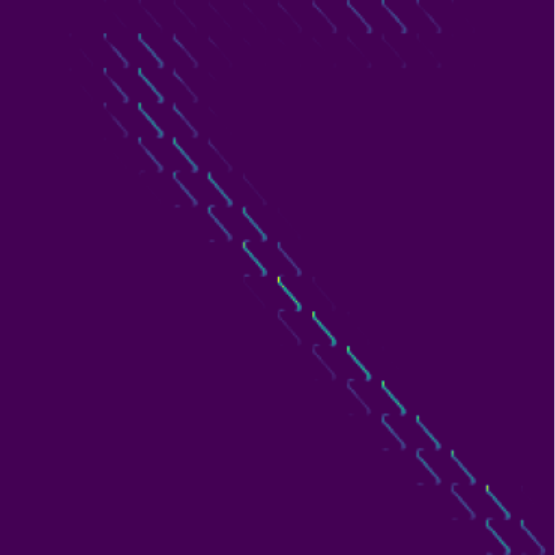} 
%\vspace{-1.5\baselineskip}
\end{subfigure} 
\begin{subfigure}{0.19\textwidth}
\includegraphics[height=\linewidth]{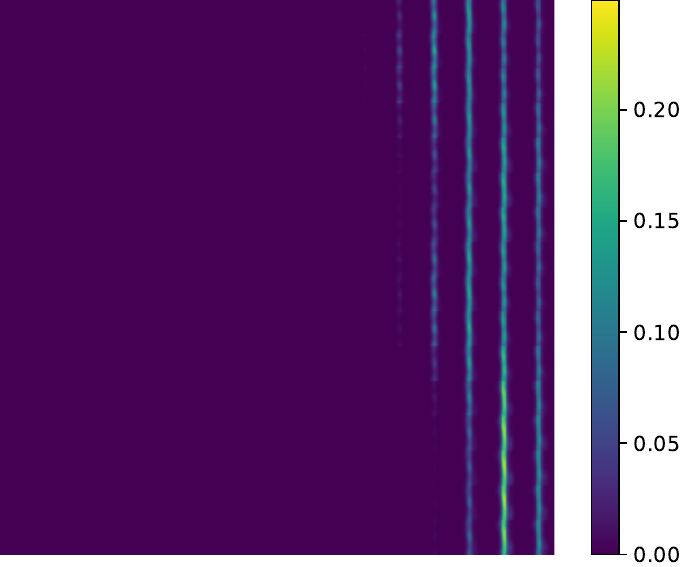} 
%\vspace{-1.5\baselineskip}
\end{subfigure}

\caption{Visualization of $\Q$ and $\K$ with random initialization. 
The upper section of the figure is initialization (before training), while the lower section is trained. 
In each section, the first row is $\Q\K$, the second row is $\P\Q\K\P^{T}$, and the third row is $\mbox{SoftMax}(\P\Q\K\P^{T})$. 
Columns are laid out as layers 1, 3, and 6 (depth is 6). 
Because the 8 heads in each layer have similar patterns, we only choose head 3 for illustration for all the figures.}
\label{fig:random_init}
\end{figure*}

Visualization of the weights and attention maps with random, mimetic~\cite{trockman2023mimetic}, and impulse initialization before and after training can be found in~\cref{fig:impulse_init},~\cref{fig:mimetic_init} and~\cref{fig:random_init}. 
It is worth noting that the positional encoding used here is the concatenation of $\mbox{sin}$ and $\mbox{cos}$ of $x$ and $y$ coordinates separately. 
Consequently, at times in $\Q\K$, some peculiar patterns may appear in $4\times 4$ blocks. 
However, the real attention map still requires multiplication by the positional encoding on both sides $\P\Q\K\P^{T}$.

% To split the supplementary pages from the main paper, you can use \href{https://support.apple.com/en-ca/guide/preview/prvw11793/mac#:~:text=Delete%20a%20page%20from%20a,or%20choose%20Edit%20%3E%20Delete).}{Preview (on macOS)}, \href{https://www.adobe.com/acrobat/how-to/delete-pages-from-pdf.html#:~:text=Choose%20%E2%80%9CTools%E2%80%9D%20%3E%20%E2%80%9COrganize,or%20pages%20from%20the%20file.}{Adobe Acrobat} (on all OSs), as well as \href{https://superuser.com/questions/517986/is-it-possible-to-delete-some-pages-of-a-pdf-document}{command line tools}.

{
    \small
    \bibliographystyle{ieeenat_fullname}
    \bibliography{main}
}

% WARNING: do not forget to delete the supplementary pages from your submission 
% \input{sec/X_suppl}

\end{document}